\documentclass[letterpaper, 10 pt, conference]{ieeeconf}  %
\usepackage[letterpaper, left=48pt, right=60pt, bottom=58pt, top=55pt]{geometry}

\usepackage{caption}

\makeatletter
\let\NAT@parse\undefined
\makeatother

\usepackage{xcolor}
\definecolor{gold}{HTML}{daaa00}
\usepackage{amssymb, amsmath, mathtools}
\usepackage{graphicx}
\usepackage{float}
\usepackage{multirow}
\usepackage{booktabs}
\usepackage{svg}
\usepackage{subcaption}
\usepackage[utf8]{inputenc}
\usepackage[T1]{fontenc}
\usepackage[english]{babel}
\usepackage{svg}
\usepackage{overpic}
\usepackage{algorithm}
\usepackage{algpseudocode}
\usepackage[colorlinks, citecolor=gold, linkcolor=gold, urlcolor=gold]{hyperref}

\DeclareMathOperator*{\argmin}{arg\,min}

\makeatletter
\let\NAT@parse\undefined
\makeatother
\usepackage[numbers,sort&compress]{natbib}
\usepackage[nameinlink,capitalise]{cleveref}

\IEEEoverridecommandlockouts                              %

\newcommand{\grenderwidth}{0.137\textwidth}

\title{\LARGE \bf
VLAD-Grasp: Zero-shot Grasp Detection via Vision-Language Models
}

\author{Manav Kulshrestha$^{1}$, S. Talha Bukhari$^{1}$, Damon Conover$^{2}$, Aniket Bera$^{1}$%
\thanks{\raggedright $^{1}$Manav Kulshrestha, S. Talha Bukhari, and Aniket Bera are with the IDEAS Lab, Department of Computer Science, Purdue University, West Lafayette, IN, USA. %
        {\tt\small \{\href{mailto:mkulshre@purdue.edu}{mkulshre}\allowbreak,\href{mailto:bukhars@purdue.edu}{bukhars}\allowbreak,\href{mailto:aniketbera@purdue.edu}{aniketbera}\allowbreak\}@\allowbreak purdue\allowbreak.edu}}
\thanks{\raggedright $^{2}$Damon Conover is with the DEVCOM Army Research Laboratory, Adelphi, MD, USA. %
          {\tt\small \href{mailto:damon.m.conover.civ@army.mil}{damon.m.conover.civ@army.mil}}}%
}

\begin{document}

\hypersetup{urlcolor=black}
\maketitle
\hypersetup{urlcolor=gold}
\thispagestyle{empty}
\pagestyle{empty}

\begin{abstract}
Robotic grasping is a fundamental capability for enabling autonomous manipulation, with usually infinite solutions.
State-of-the-art approaches for grasping rely on learning from large-scale datasets comprising expert annotations of feasible grasps.
Curating such datasets is challenging, and hence, learning-based methods are limited by the solution coverage of the dataset, and require retraining to handle novel objects.
Towards this, we present VLAD-Grasp, a \underline{V}ision-\underline{L}anguage model \underline{A}ssisted zero-shot approach for \underline{D}etecting \underline{Grasp}s. 
Our method
(1) prompts a large vision-language model to generate a goal image where a virtual cylindrical proxy intersects the object's geometry, explicitly encoding an antipodal grasp axis in image space, then
(2) predicts depth and segmentation to lift this generated image into 3D, and
(3) aligns generated and observed object point clouds via principal components and correspondence-free optimization to recover an executable grasp pose.
Unlike prior work, our approach is training-free and does not require curated grasp datasets, while achieving performance competitive with the state-of-the-art methods on the Cornell and Jacquard datasets.
Furthermore, we demonstrate zero-shot generalization to real-world objects on a Franka Research 3 robot, highlighting vision-language models as powerful priors for robotic manipulation.
\end{abstract}

\section{Introduction}\label{sec:intro}

Grasp detection entails computing a pose for the robotic gripper, such that closing the gripper fingers results in a stable grasp on the object, allowing downstream robotic manipulation tasks, such as rearrangement~\cite{kulshrestha2023structural, zhai2024sg}, sorting~\cite{koskinopoulou2021robotic}, assembly~\cite{zhang2024residual}, and human–robot collaboration~\cite{keshari2023cograsp,chen2025multimodal}.
Hence, grasp detection is a fundamental capability required for robot autonomy in unstructured environments.
However, identifying stable grasps is challenging due to the diversity of object geometries, material properties, and task constraints, as well as the nature of the task, usually comprising infinite feasible solutions.
Therefore, common-sense reasoning capabilities are a necessary precursor to robust grasp detection.

Classical approaches to grasp detection rely on analytical models of contact mechanics and object geometry for predicting grasp stability~\cite{bicchi2000robotic, murray2017mathematical}.
These methods provide useful insights, but are limited by simplifying heuristics, which restrict their ability to generalize across arbitrary shapes and materials.
In recent years, learning-based methods have provided much more feasible solutions to grasp detection, enabled by large-scale datasets comprising expert grasp annotations \cite{newbury2023deep}.
Deep networks trained on annotated grasps \cite{depierre2018jacquard} or simulation-based labels \cite{jiang2011efficient} achieve high accuracy in distribution.
However, their performance is upper-bounded by the coverage and quality of the training datasets.
Collecting expert-annotated data is costly and time-intensive, and simulated grasps are often biased toward restricted families of grasp modes \cite{li2024jacquard}.
As a result, most supervised methods fail to generalize to novel object categories, cluttered environments, and real-world deployment scenarios.
Recent work has attempted to address this gap by incorporating semantic or multi-modal cues into grasp prediction, such as via language-conditioning~\cite{shridhar2022cliport, xu2023joint, vuong2024language} and segmentation-guidance~\cite{noh2025graspsam}.
While such approaches improve task relevance, they still depend on retraining with annotated grasp data and, therefore, inherit similar generalization and scalability issues.

\begin{figure}[t]
    \centering\vspace{-1.5mm}
    \includegraphics[width=\linewidth,trim=0 6.3cm 0 0,clip]{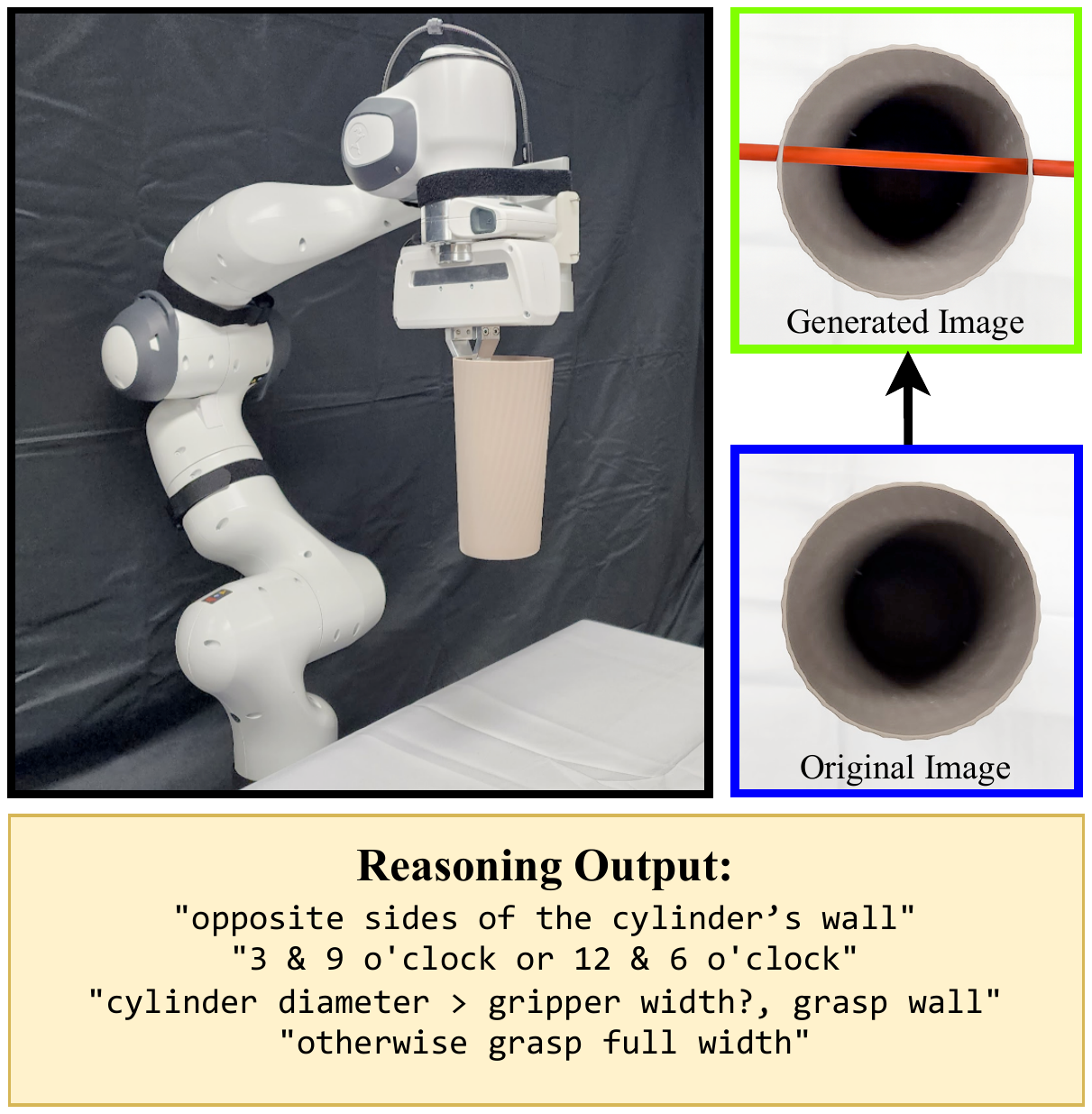}
    \caption{From an RGB-D image of the object, we query a VLM with sequential guiding prompts to reason about the object’s geometry and grasp feasibility.
    Next, the VLM generates a goal image depicting a virtual cylindrical proxy intersecting the object, which encodes the antipodal grasp axis in image space.
    This axis is lifted in 3D and aligned with the scene observation to yield an executable grasp pose.
    }
    \label{fig:summary}
    \vspace{-4mm}
\end{figure}

In this work, we ask a different question:
Can the general-purpose reasoning and generative capabilities of large vision–language models (VLMs), trained on internet-scale multi-modal data, be directly exploited for robotic grasping \emph{without any task-specific training or fine-tuning}?
VLMs encode broad visual–semantic knowledge from massive image–text corpora, including implicit understanding of object affordances and human–object interactions \cite{hurst2024gpt, comanici2025gemini, team2024chameleon}.
This makes them a compelling prior for robotic grasping, where explicit supervision is costly and incomplete.

Towards this, we present VLAD-Grasp, a zero-shot grasp synthesis framework that leverages large pretrained VLMs to detect feasible grasps on unseen objects.
A single RGB-D observation of the scene is used to query a VLM, along with a sequence of guiding textual prompts to generate a goal image containing a virtual cylindrical proxy intersecting the object geometry, encoding the antipodal grasp axis in image space.
Next, monocular depth estimation and segmentation lift the generated image into 3D, followed by alignment with original object point clouds via principal component analysis (PCA) and correspondence-free point cloud optimization, yielding an executable grasp pose for the robot.
Crucially, our approach does not require any expert grasp datasets, supervision, or fine-tuning, bypassing the limitations of supervised methods trained on expert grasp annotations, while maintaining competitive performance.

The novelty of VLAD-Grasp lies in combining three elements:
\begin{enumerate}
    \item \emph{Zero-shot grasping} via VLM prompts, which reasons about grasp feasibility and spatial relationships by leveraging its large-scale vision-language pretraining, removing the need for curated grasp annotations.
    \item \emph{Geometric consistency alignment}, which combines monocular depth prediction, principle component-based registration, and correspondence-free point cloud optimization to consistently recover grasp poses from generated grasp proposals in image space.
    \item \emph{Open vocabulary grasping}, which transfers to grasping objects of arbitrary geometry and appearance without retraining.
\end{enumerate}

\noindent We provide extensive evaluations on the Cornell and Jacquard datasets, demonstrating competitive performance against supervised baselines, and significantly outperform prior training-free approaches for grasping.

\section{Related Work}\label{sec:related}

\subsection{Large-scale Pretraining for Robotics}

Traditional data-driven methods in robotics have been trained on task-specific datasets, which are challenging to curate exhaustively for the task at hand, and the trained methods fail to generalize beyond the experimental setting.
Recent years have seen a push for large-scale data collection and pre-training to solve robotics tasks \cite{o2024open,firoozi2025foundation}, inspired by the success of large language models trained on web-scale datasets \cite{achiam2023gpt, touvron2023llama}.
The emergent behaviors from large-scale pretraining include common-sense reasoning across various modalities, such as text, images, audio, and video.
\citet{kapelyukh2023dall} demonstrate human-like object rearrangement skills using diffusion models trained on web-scale data.
\citet{shridhar2022cliport} utilize CLIP for semantic reasoning and Transporter \& Perceiver for spatial reasoning.
\citet{karamcheti2023language} presents a framework for language-driven representation learning in robotics by capturing semantic, spatial, and temporal representations learned from videos and captions.

In this work, we propose an approach that leverages the visual and textual reasoning capabilities of large vision-language models to infer grasping from 2D images.
We solely used the VLM's general purpose reasoning capabilities to infer spatial and semantic relationships between geometries to detect feasible grasps in image space, without requiring explicit training on expert data.

\subsection{Grasp Detection}

Detecting grasp candidates is a core task in robotic manipulation.
Classical methods use geometric approaches to generate and evaluate stable grasps \cite{bicchi2000robotic, murray2017mathematical, newbury2023deep}.
With the advent of modern deep learning, a wide range of approaches for grasp generation has been developed.
\citet{morrison2018closing, kumra2020antipodal, ainetter2021end} train convolutional neural networks to detect grasps from RGB-D images.
\citet{mahler2017dex, mousavian20196, bukhari2025variational} infer stable grasp distributions in 3D by ingesting object point clouds.
\citet{shridhar2022cliport, xu2023joint, vuong2024language} leverage language information to guide grasp synthesis via text prompts.
\citet{noh2025graspsam} incorporate large-scale pretraining of the Segment-Anything Model (SAM) to guide grasp detection via object masks.

However, all the aforementioned methods involve training on expert grasp data, which is time-consuming to collect and tedious to train reliably on \cite{eppner2021acronym, depierre2018jacquard, vuong2024grasp}.
Furthermore, such datasets may sometimes miss crucial intuitive grasp candidates that the employed expert was unable to model \cite{li2024jacquard}.
This limits the performance of trained grasp detection methods, without incorporating intuitive reasoning.
ShapeGrasp \cite{li2024shapegrasp} proposes a training-free approach to detect grasps by decomposing an object into its constituent geometries, and then converting the decomposition into an object-graph which is fed to an LLM to reason about their spatial relationships.
However, the compression of visual features into textual prompts that the LLM can parse discards the rich visual context needed to generate stable grasps over arbitrarily complex geometries.
Towards this objective, we propose using the reasoning capabilities of large pretrained VLMs to generate grasp candidates directly from images, without requiring any training on expert data or explicit geometric decomposition.

In a different vein, prior works exist which use language models to reason over a set of grasp candidates to infer a feasible subset that is the most promising \cite{tang2023graspgpt, tang2025foundationgrasp}.
Such methods still require supervised training to yield a grasp evaluator (classifier), and are coupled with a grasp generator to generate the grasp candidates.
Although we focus on grasp generation in this work, our approach internally performs grasp reasoning in a similar spirit to grasp evaluation methods.
Furthermore, our method can be integrated with grasp evaluators in a plug-and-play fashion.

\section{Methodology}\label{sec:method}

\begin{figure*}
    \centering
    \includegraphics[width=0.99\linewidth,trim={0.9cm, 0.4cm, 0.7cm, 0.2cm}, clip]{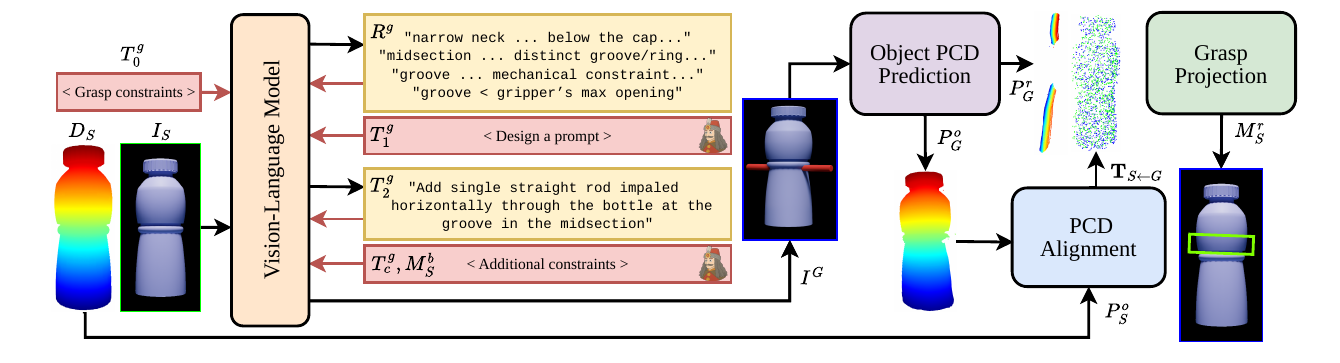}
    \caption{Overview of our approach. We capture an RGB-D image $(I_S, D_S)$ of the object and mask out background distractors. The RGB image $I_S$ is provided to the VLM, following structured guiding prompts $T^g_i$, to help it reason $R^g$ about object geometry and eventually produce a generated image $I_G$, where the goal grasp is indicated by a rod passing through the antipodal grasp points on the object's surface. Following this, predict a point cloud $P^o_G$ for the object in the generated image $I_G$ and match it with the point cloud $P^o_S$ for the object in the original image $I_S$
    }\label{fig:pipeline}
    \vspace{-2mm}
\end{figure*}

We tackle the problem of generating a grasp from a single RGB-D image of an object in a zero-shot manner, for which our end-to-end approach is detailed in Fig.~\ref{fig:pipeline}.
Our approach comprises three major components:
(1) Querying a VLM with an image of an object, along with grasp constraints, to generate another image comprising a gripper proxy demonstrating a grasp on the object,
(2) lifting the objects into 3D space using monocular depth estimates to align their representations, and
(3) projecting the grasp proposal onto the input image space.

\setlength{\textfloatsep}{6pt}
\begin{algorithm}[t]
    \footnotesize
    \caption{VLAD-Grasp: Zero-Shot Grasp Detection}
    \label{alg:vlad_grasp}
    \begin{algorithmic}[1]
        \Require RGB-D image $(I_S, D_S)$, intrinsics $K$, extrinsics $T$,
                 prompts $T^g_0, T^g_1, T^g_c$, gripper params $w^g_S, h^g_S$,
                 compliance $\lambda$, threshold $\delta$
        \Ensure Grasp rectangles $\mathcal{G}$
        
        \Statex \textbf{// Stage 1: Grasp Image Generation}
        \State $M^o_S \leftarrow \textsc{Segment}(I_S)$
            \Comment{Object mask}
        \State $M^b_S \leftarrow \mathbf{1} - M^o_S$
            \Comment{Background mask}
        \State $R^g \sim p_\theta(\,\cdot\,|I_S, T^g_0)$
            \Comment{Geometric reasoning}
        \State $T^g_2 \sim p_\theta(\,\cdot\,|R^g, T^g_1)$
            \Comment{Image-editing directive}
        \State $I_G \sim p_\theta(\,\cdot\,|T^g_c, T^g_2, M^b_S, I_S)$
            \Comment{Goal image with rod $r_G$}
        
        \Statex \textbf{// Stage 2: Object Alignment}
        \State $M^o_G, M^r_G \leftarrow \textsc{Segment}(I_G)$
            \Comment{Object and rod masks}
        \State $D_G \leftarrow \textsc{DepthEstimate}(I_G)$
            \Comment{Monocular depth}
        \State $P^o_S \leftarrow f_{K,T}(D_S, M^o_S)$
            \Comment{Scene object PCD}
        \State $P^o_G \leftarrow f_{K,T}(M^o_G, D_G)$
            \Comment{Goal object PCD}
        \State $P^r_G \leftarrow f_{K,T}(M^r_G, D_G)$
            \Comment{Goal rod PCD}
        \State $\mathbf{V}^o_S, \mathbf{\Lambda}^o_S \leftarrow \textsc{PCA}(P^o_S)$
        \State $\mathbf{V}^o_G, \mathbf{\Lambda}^o_G \leftarrow \textsc{PCA}(P^o_G)$
            \Comment{Eigenvectors and eigenvalues}
        \State $\mathbf{R}^*_{S \leftarrow G} \leftarrow
            \textsc{CfOpt}(\mathbf{V}^o_S, \mathbf{\Lambda}^o_S, \mathbf{V}^o_G, \mathbf{\Lambda}^o_G, P^o_S, P^o_G)$
            \Comment{Correspondence-free opt.}
        \State $\mathbf{T}_{S \leftarrow G} \leftarrow
            \textsc{RigidTransform}(\mathbf{R}^*_{S \leftarrow G}, P^o_S, P^o_G)$
            \Comment{Optimal rigid transform}
        
        \Statex \textbf{// Stage 3: Grasp Projection}
        \State $P^r_S \leftarrow P^r_G\, \mathbf{T}^\top_{S \leftarrow G}$
            \Comment{Rod in scene frame}
        \State $M^r_S \leftarrow \pi_1(f^{-1}_{K,T}(P^r_S))$
            \Comment{Project rod to image}
        \State ${c}^r, {u}^r \leftarrow \textsc{FitLine}(M^r_S)$
            \Comment{Rod centroid and axis}
        \State $y \leftarrow \textsc{Discontinuities}(M^r_S, {c}^r, {u}^r)$
            \Comment{Rod discontinuities}
        \State $\tilde{y} \leftarrow \{(s_i, l_i) \in y \mid \delta \leq l_i \leq w^g_S\}$
            \Comment{Filter valid discontinuities}
        \State $\mathcal{G} \leftarrow \{(\mathbf{c}^r,\ \lambda l_i,\ h^g_S) \mid (s_i, l_i) \in \tilde{y}\}$
            \Comment{Grasp rectangles}
        \State \Return $\mathcal{G}$
    \end{algorithmic}
\end{algorithm}

\subsection{Grasp Image Generation}\label{subsec:gen}

State-of-the-art VLMs learn to auto-regressively model an implicit joint linguistic–visual distribution $p_\theta(\mathcal T, \mathcal I)$.
This can later be conditioned on a sequence of elements $(x_k)^{t-1}_{k=1} \subseteq (\mathcal T \cup \mathcal I)^{t-1}$ belonging to either modality, to generate the next token $x_t \sim p_\theta(\,\cdot\,|x_{<t})$ where $x_t \in \mathcal T \cup \mathcal I$.
We aim to leverage the strong reasoning capabilities of VLMs while capitalizing on image generation capabilities
to provide a natural medium for representing geometric information and guiding the downstream grasping task.

To achieve this, we devise a three-step prompting procedure.
First, we provide the VLM with an image $I_S$ of the object and query it with a text prompt $T_0^g$ comprising the constraints for the desired grasp:
Gripper size and configuration, and a preference for grasps whose antipodal contact points are both visible in image space.
While it is not always geometrically guaranteed that both contact points are simultaneously visible, framing this as a soft constraint encourages the VLM to favor grasps where more of the rod is exposed in the generated image $I_G$, which in turn yields a more complete rod mask $M^r_S$ and improves downstream grasp interpretation.
To minimize distractions for the VLM, we exclude the background using an object mask $M^o_S$ obtained from a segmentation model.
The VLM responds with an intermediate geometric reasoning $R^g \sim p_\theta(\,\cdot\,|I_S, T_0^g)$, which translates the abstract constraints in $T_0^g$ into what they imply for the specific object geometry in $I_S$.
Qualitatively, the model interprets visual shape cues (e.g., bottle neck vs.\ wide body, mug rim vs.\ handle, box corner vs.\ flat face) to identify where antipodal contacts could plausibly lie on this particular object, while respecting the gripper limits and ensuring both contact locations are visible in the image.
In the second step, we query the VLM with a prompt $T_1^g$ that instructs it to consolidate its geometric reasoning $R^g$ into a precise, self-contained image-editing directive $T_2^g \sim p_\theta(\,\cdot\,|R^g, T_1^g)$, specifying exactly where and how a cylindrical rod $r_G$ should be rendered impaling the object $o_G$ in $I_S$ so as to encode the antipodal grasp axis as a physical object in image space.
Separating this from the first step is important because geometric reasoning --- identifying \textit{where} to grasp --- and instruction synthesis --- specifying \textit{how} to render that grasp --- are distinct cognitive tasks that the model handles more reliably when not conflated into a single prompt~\cite{wei2022chain}.
In the third step, we augment $T_2^g$ with additional text constraints $T^g_c$ and an inpainting mask $M^b_S$ (the inverse of $M^o_S$) to penalize modifications to the object $o_G$ during generation, and query the model with this constraint-augmented prompt to obtain the goal image $I_G \sim p_\theta(\,\cdot\,|T^g_c, T^g_2, M^b_S, I_S)$ in which the rod $r_G$ intersects the object geometry at the predicted antipodal grasp axis.

Representing the grasp as a cylindrical rod has three main advantages.
First, representing the grasp as a physical object in the generated image helps ground the generation from the VLM in concepts within its training domain, rather than abstract concepts such as stability and affordances. %
Second, a straight rod implicitly encodes that the two grasp points are antipodal with respect to the object's surface, which is consistent with our experimental setup comprising a two-finger, antipodal gripper.
Third, a cylindrical rod is a primitive geometry, which prevents the focus bandwidth of the VLM from shifting from the task at hand, enabling better prompt adherence.

\subsection{Object Alignment}\label{subsec:align}

To guide the VLM's grasp image generation, we include text prompts $T^g_c$ and inpainting masks $M^b_S$ to minimize changes to the object between the original image $I_S$ and the generated image $I_G$.
However, these specifications only serve as soft constraints for current state-of-the-art VLMs for image editing.
This allows us to provide conservative constraints in the text prompt $T^g_c$ and directly use the background mask $M^b_S$ without conflicting with the model's logically consistent image generation capability in order to represent the full range of impaling rods (i.e., antipodal grasps).
Consequently, however, there are differences with respect to object representation and the overall composition between the original image $I_S$ and the generated image $I_G$, making it challenging to directly interpret the cylindrical rod $r_G$ in the generated image $I_G$ as an actionable grasp pose for the object $o_S$ in the original image $I_S$.
To address this, we align the object $o_S$ in the original image with its counterpart $o_G$ in the generated image, assuming spatial consistency within each image.
This assumption holds well in our experiments, advocating for the representation ability and contextual consistency of large VLMs.

Using a segmentation model, we obtain the masks $M^o_G,M^r_G$ for both the object $o_G$ and rod $r_G$ in the generated image $I_G$.
Let $f_{K,T}: \mathcal M  \times \mathcal D \to \mathcal P(\mathbb R^3)$ be the pinhole lifting (back-projection) operator that converts a masked depth map $(M, D) \in \mathcal M\times \mathcal D$ into the object point cloud $P \subseteq \mathbb R^3$, parametrized by the camera intrinsics $K$ and extrinsics $T$. We utilize a monocular depth estimation model to predict depth image $D_G$ for the generated image $I_G$ and combine it with the masks $M^o_G,M^r_G$ to obtain estimated point clouds $P^o_G = f_{K,T}(M^o_G, D_G) \subseteq \mathbb R^3$ and $ P^r_G = f_{K,T}(M^r_G, D_G) \subseteq \mathbb R^3$ for elements $o_G, r_G$. 
Note that we already have the object mask $M^o_S$ and true depth $D_S$ for the original image and, therefore, use it to also construct a point cloud $P^o_S = f_{K,T}(D_S, M^o_S) \subseteq \mathbb R^3$ for the object in the original scene.
Following this, we can perform object alignment between objects $o_S,o_G$ using their point clouds $P^o_S,P^o_G$.
A key observation here is that while the object in the generated image may differ in pose and details, the overall shape for $o_S$ and $o_G$ remains largely consistent. 
Therefore, we can exploit this property to perform point cloud registration that is invariant to local features, but mindful of global shape.
We mean-center both object point clouds $P^o_S,P^o_G$, estimate their covariance matrices $\mathbf\Sigma^o_S, \mathbf\Sigma^o_G$:
\[\mathbf \Sigma^o_X = \frac{1}{k-1}\left(\tilde{P^o_X}\right)^\top \tilde{P^o_X}\text{ and }\tilde{P^o_X}=P^o_X-\frac1k\mathbf 1_k^\top P^o_X\]
where $\text{rows}(P^o_X)=k$, and perform eigen decomposition 
$\mathbf\Sigma_X^o \approx \mathbf V^o_X\mathbf\Lambda_X^o\mathbf (\mathbf V^o_X)^{-1}$ to obtain for each point
cloud $P^o_S,P^o_G$, the top 3 eigenvectors $\mathbf V^o_S,\mathbf V^o_G$ (associated with the largest three eigenvalues $\mathbf\Lambda^{o}_S,
\mathbf\Lambda^{o}_G$). 
Following this, our optimal scaled rotation
becomes the following:
\[\mathbf R^*_{S\leftarrow G} = \argmin_{\mathbf {R_A}} \mathcal L\{P^o_S, P^o_G(\mathbf {R_A})^\top\}\]
where $\mathbf A = \text{diag}(i,j,k)$ and $(i,j,k) \in \{-1,1\}^3$.
$\mathcal L$ is the correspondence-free loss metric (we use Chamfer Distance~\cite{barrow1977parametric}), and $\mathbf{R_A}$ is a candidate alignment rotation given by:
\[
\mathbf R_\mathbf{A} = \mathbf V^o_S\left(\mathbf A \mathbf V^o_G\sqrt{\mathbf \Lambda^o_G}\left(\mathbf \Lambda_S^o\right)^{-1}\right)^{-1}
\]
Consequently, we get the optimal transform $\mathbf T_{S\leftarrow G}$ that aligns $P^o_G$ with $P^o_S$ by minimizing $\mathcal{L}\{P^o_S, \mathbf T_{S\leftarrow G}P^o_G\}$:
\[
\mathbf T_{S\leftarrow G} = 
\begin{bmatrix}
    \mathbf R^*_{S\leftarrow G} & \frac1m\mathbf1_n^\top P_S^o - \frac1n\mathbf R^*_{S\leftarrow G}\mathbf1_n^\top P^o_G\\
    \mathbf 0^\top & 1
\end{bmatrix}
\]
where $\text{rows}(P^o_S)=m$ and $\text{rows}(P^o_G)=n$.

\subsection{Grasp Projection}\label{subsec:grasp}

We now apply the transform to the rod point cloud $P^r_G$ to obtain its image $P^r_S = {P^r_G}\,\mathbf T_{S\leftarrow G}^\top$ in the original scene.
At this stage, the rod point cloud $P^r_S$ \emph{impales} the object point cloud $P^o_S$, indicating a set of 6-DoF grasp poses (varying in the grasp approach vector, orthogonal to the rod axis) in the original scene.
For our experimental setup, we project the rod point cloud $P^r_S$ down to a rod mask $M^r_S = \pi_1(f_{K,T}^{-1}(P^r_S))$ in the original image space, where $\pi_1: \mathcal M\times \mathcal D \to \mathcal M$ is the canonical projection which maps $\mathcal M \times \mathcal D$ onto $\mathcal M$ and returns the first component.
Hence, the projection operation naturally specifies the grasp approach vector.

For obtaining grasp rectangles (as defined in~\cite{jiang2011efficient}) from the rod mask $M^r_S$, we observe that any discontinuities in the rod $r_G$ in the generated image $I_G$, and consequently, the rod mask $M^r_S$ must be a viable grasp location since those are where the rod passes through the object.
We fit a line to the rod mask $M^r_G$ and parametrize it by the centroid of the rod mask $c^r$ and a unit vector $u$ along the largest component of the rod mask, both in the image space. Formally, let a discontinuity $y_i = (s_i, l_i) \in \mathbb R^3$ in the mask $M^r_G$ is defined by its start $s_i v^r+c^r$ and end $(s_i+l_i)u^r$ where $l_i$ is its run length. After collecting all discontinuities, we filter $\tilde y = \{(s_i, l_i)\in y\,|\,\delta\leqslant l_i \leqslant w^g_S\}$ where $\delta$ is a selection threshold and $w^g_S$ is the max width of the gripper in image space. In practice, we select $\delta$ to be the largest difference $l_k - l_{k-1}$ between successive run lengths, when sorted.
Finally, this gives us a set of valid grasps:
\[
\mathcal G = \left\{\left(
c^r,
\lambda l_i, 
h^g_S
\right)\right\}\,|\,(s_i,e_i,l_i) \in \tilde y\}
\] where, in the original image space $S$, $c^r$ is the center of the grasp, $l_i$ is the grasp width, $\lambda$ is a compliance factor, and $h^g_S$ is the height of the gripper.

\section{Experiments}

\begin{figure*}[t]
    \centering
    \begin{subfigure}{\grenderwidth}
        \caption{GR-ConvNet}
    \end{subfigure}
    \hfill
    \begin{subfigure}{\grenderwidth}
        \caption{GG-CNN}
    \end{subfigure}
    \hfill
    \begin{subfigure}{\grenderwidth}
        \caption{SE-ResUNet}
    \end{subfigure}
    \hfill
    \begin{subfigure}{\grenderwidth}
        \caption{GraspSAM}
    \end{subfigure}
    \hfill
    \begin{subfigure}{\grenderwidth}
        \caption{LGD w/ Query}
    \end{subfigure}
    \hfill
    \begin{subfigure}{\grenderwidth}
        \caption{ShapeGrasp}
    \end{subfigure}
    \hfill
    \begin{subfigure}{\grenderwidth}
        \caption{\textbf{VLAD-Grasp}}
    \end{subfigure}
    \newline
    \begin{subfigure}{\grenderwidth}
        \begin{overpic}[width=\textwidth,trim=250 250 170 230,clip]{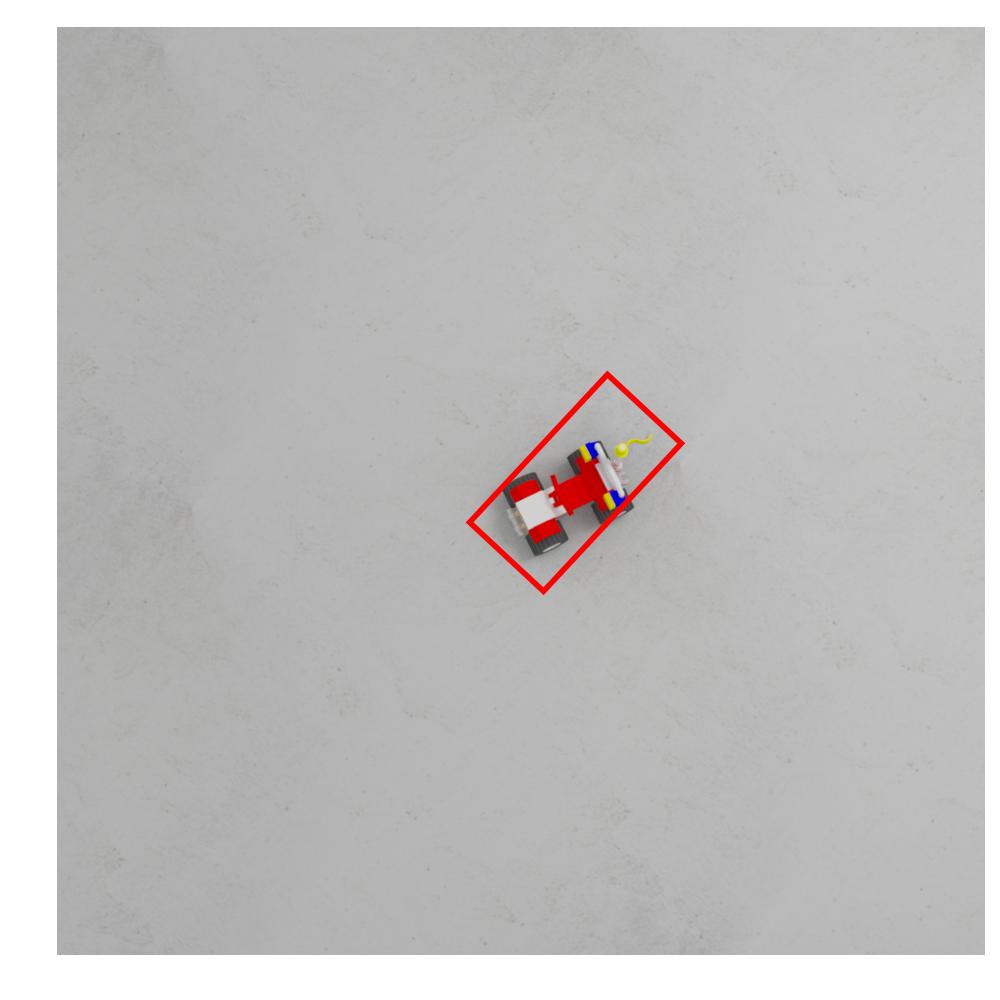}
        \end{overpic}
    \end{subfigure}
    \hfill
    \begin{subfigure}{\grenderwidth}
        \begin{overpic}[width=\textwidth,trim=250 250 170 230,clip]{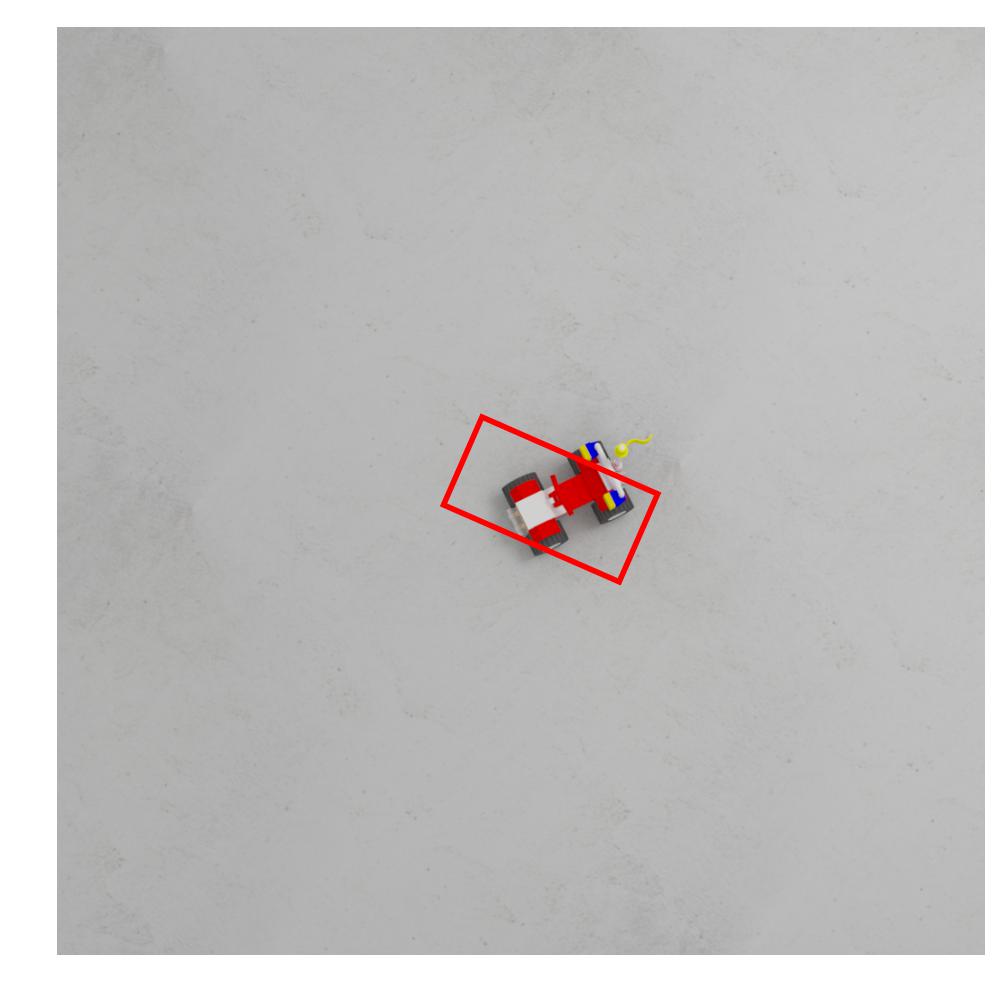}
        \end{overpic}
    \end{subfigure}
    \hfill
    \begin{subfigure}{\grenderwidth}
        \begin{overpic}[width=\textwidth,trim=250 250 170 230,clip]{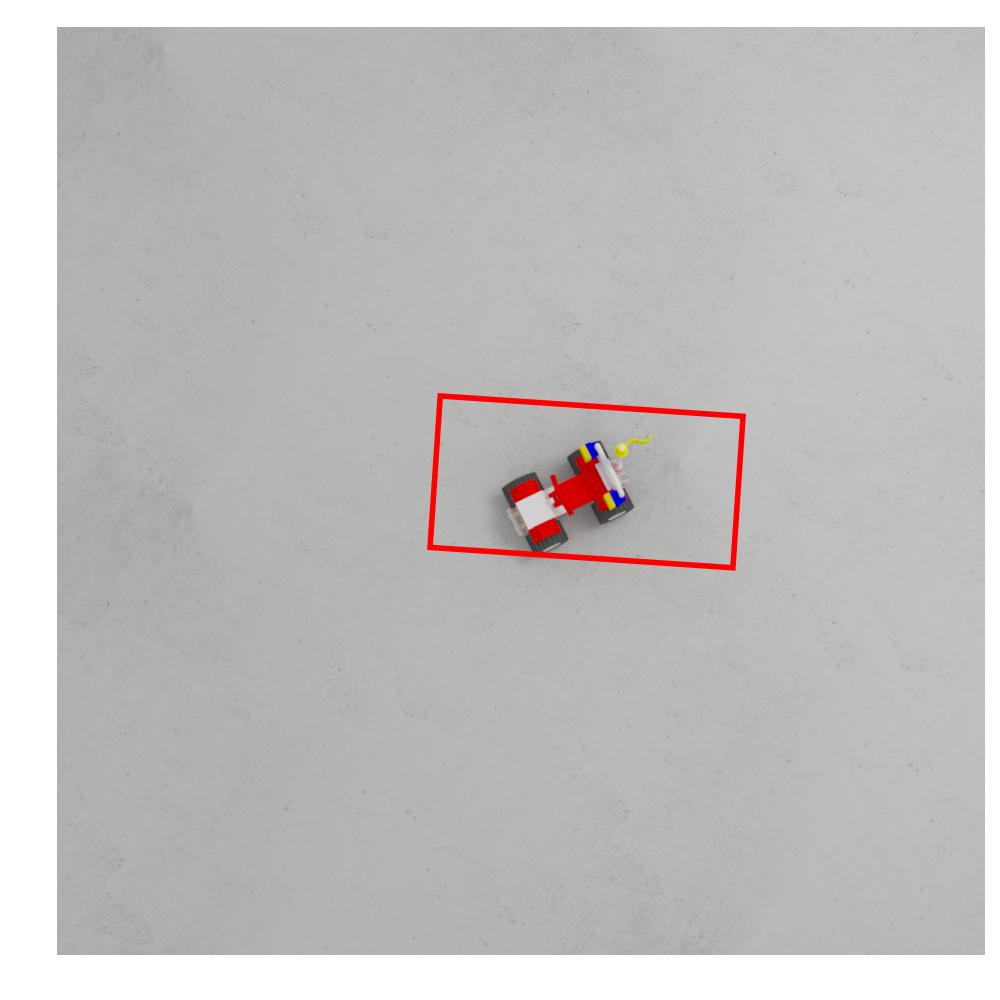}
        \end{overpic}
    \end{subfigure}
    \hfill
    \begin{subfigure}{\grenderwidth}
        \begin{overpic}[width=\textwidth,trim=250 250 170 230,clip]{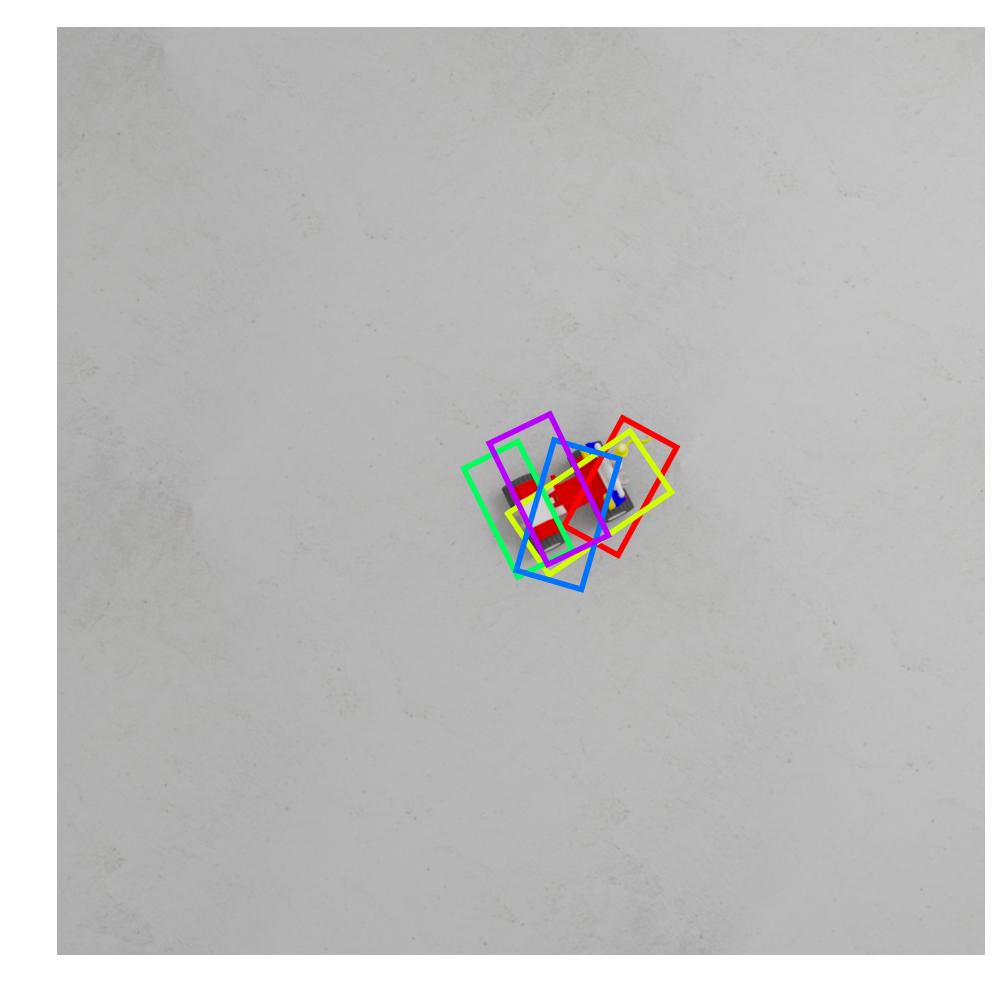}
        \end{overpic}
    \end{subfigure}
    \hfill
    \begin{subfigure}{\grenderwidth}
        \begin{overpic}[width=\textwidth,trim=250 250 170 230,clip]{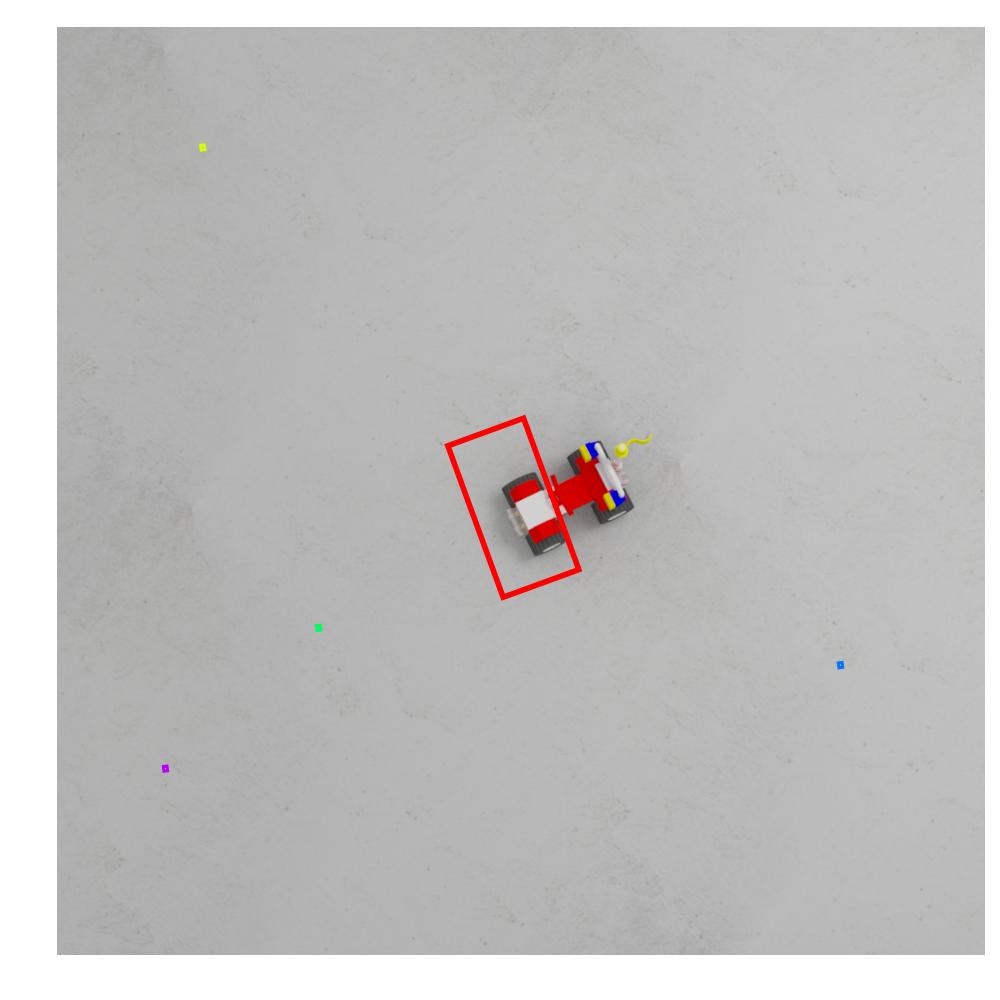}
        \end{overpic}
    \end{subfigure}
    \hfill
    \begin{subfigure}{\grenderwidth}
        \begin{overpic}[width=\textwidth,trim=250 250 170 230,clip]{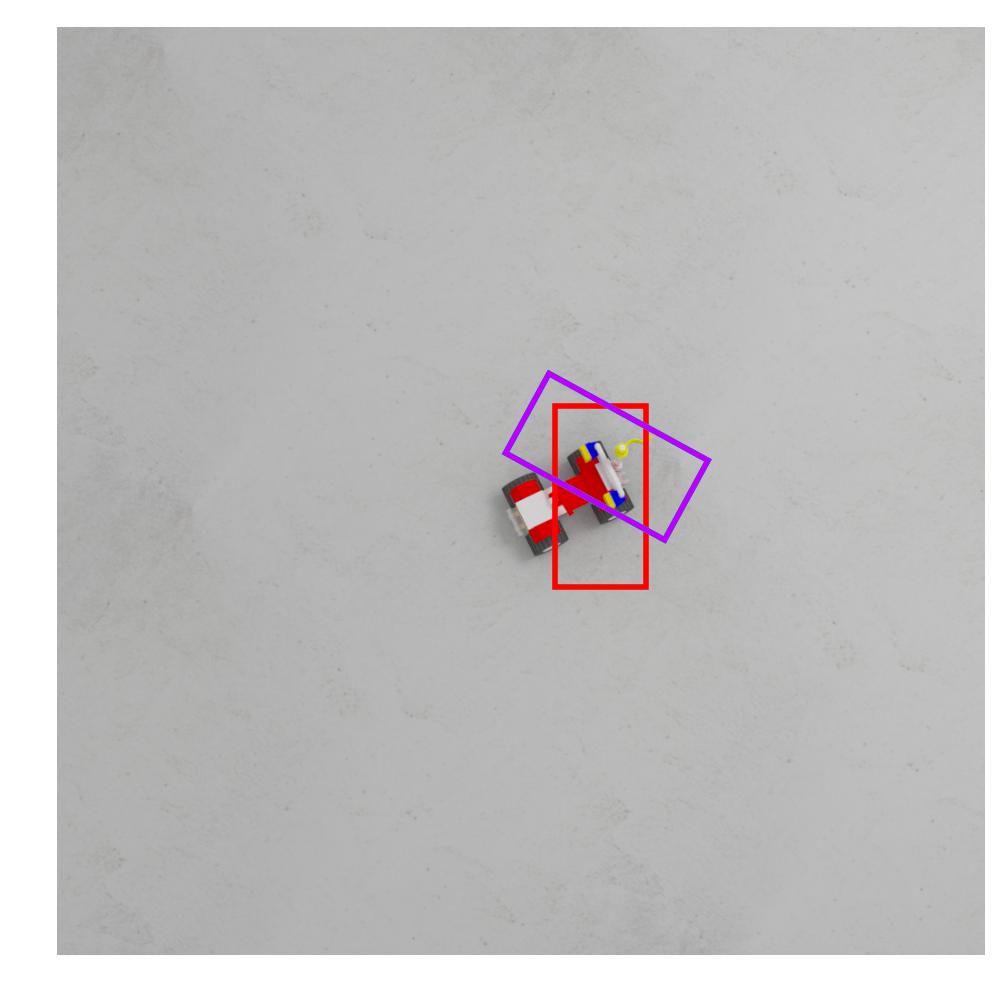}
        \end{overpic}
    \end{subfigure}
    \hfill
    \begin{subfigure}{\grenderwidth}
        \begin{overpic}[width=\textwidth,trim=250 250 170 230,clip]{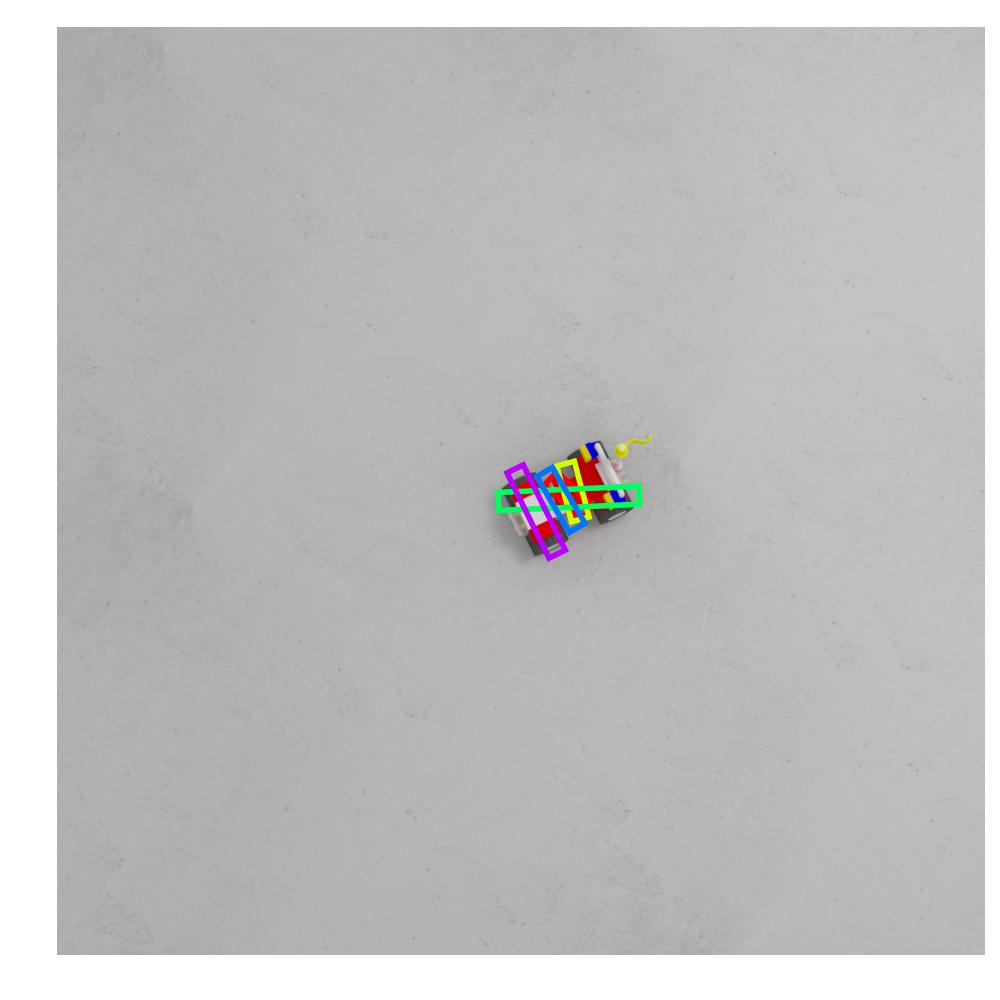}
        \end{overpic}
    \end{subfigure}
    \newline
    \begin{subfigure}{\grenderwidth}
        \begin{overpic}[width=\textwidth,trim=100 10 10 0,clip]{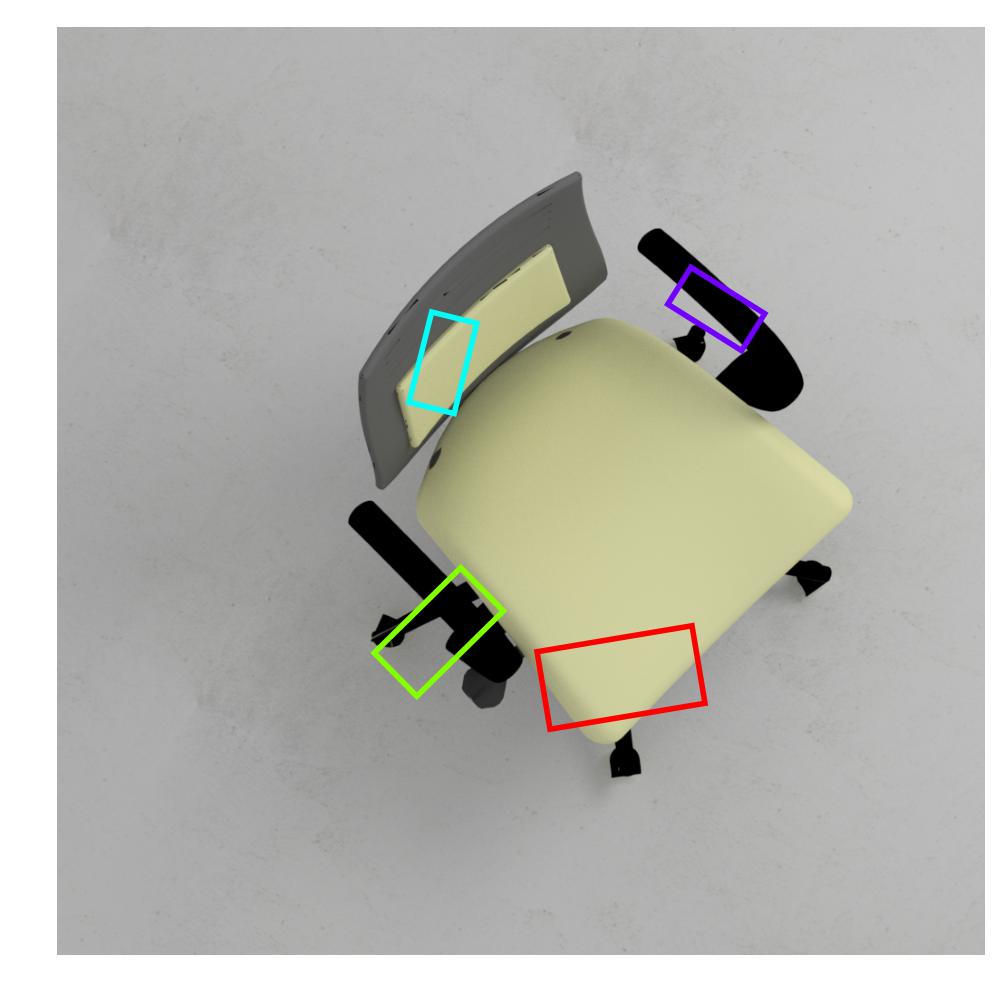}
        \end{overpic}
    \end{subfigure}
    \hfill
    \begin{subfigure}{\grenderwidth}
        \begin{overpic}[width=\textwidth,trim=100 10 10 0,clip]{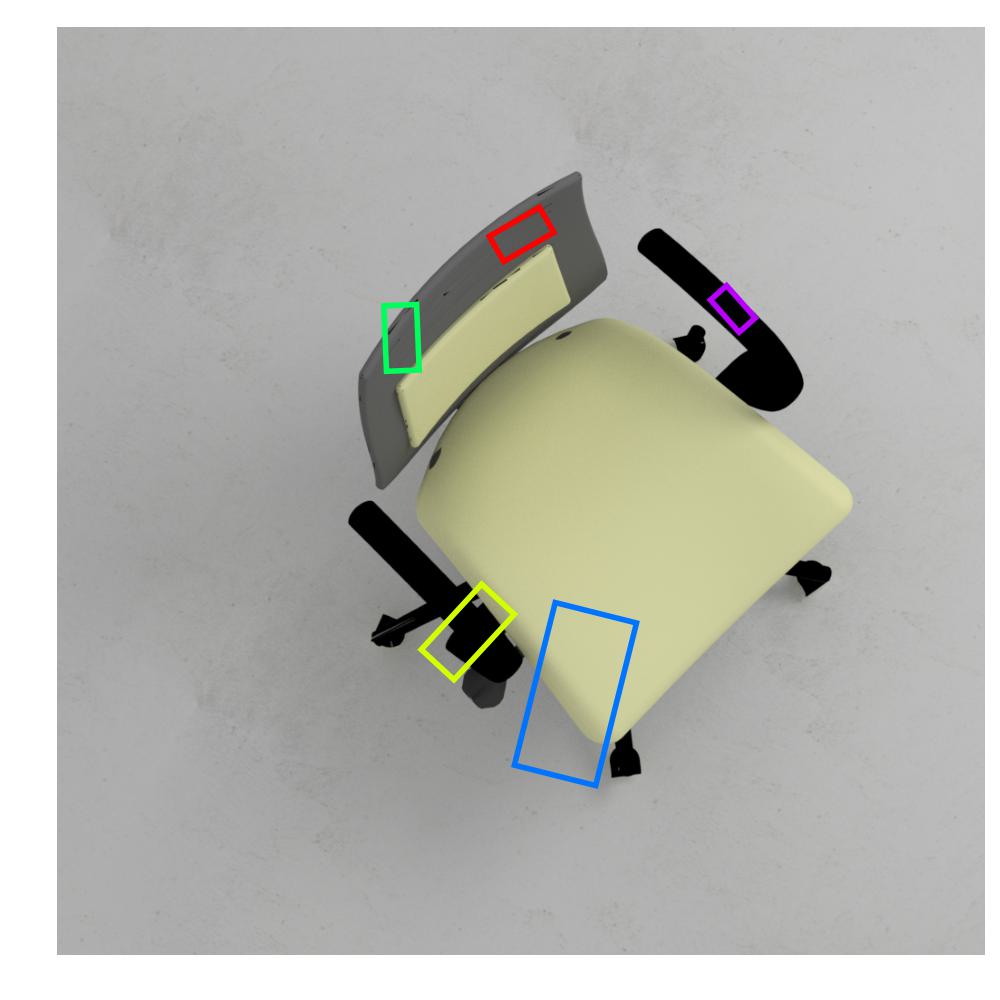}
        \end{overpic}
    \end{subfigure}
    \hfill
    \begin{subfigure}{\grenderwidth}
        \begin{overpic}[width=\textwidth,trim=100 10 10 0,clip]{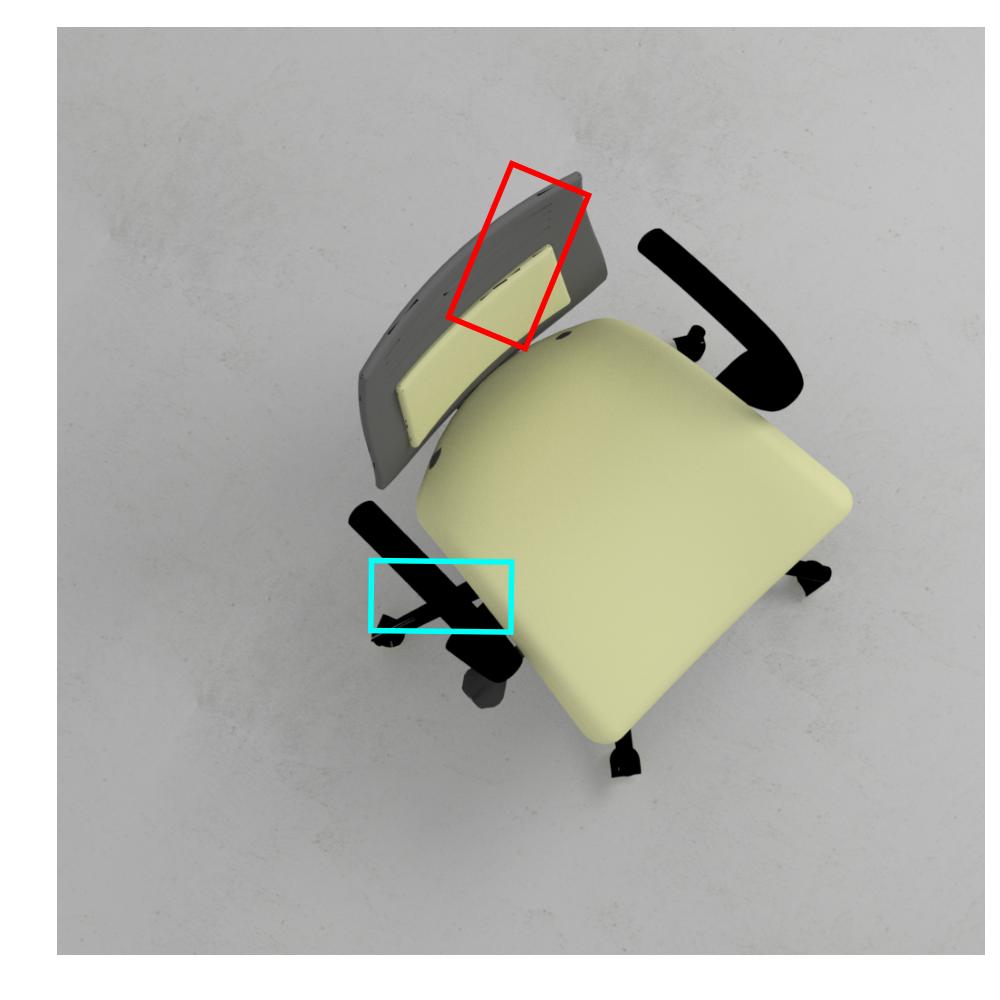}
        \end{overpic}
    \end{subfigure}
    \hfill
    \begin{subfigure}{\grenderwidth}
        \begin{overpic}[width=\textwidth,trim=100 10 10 0,clip]{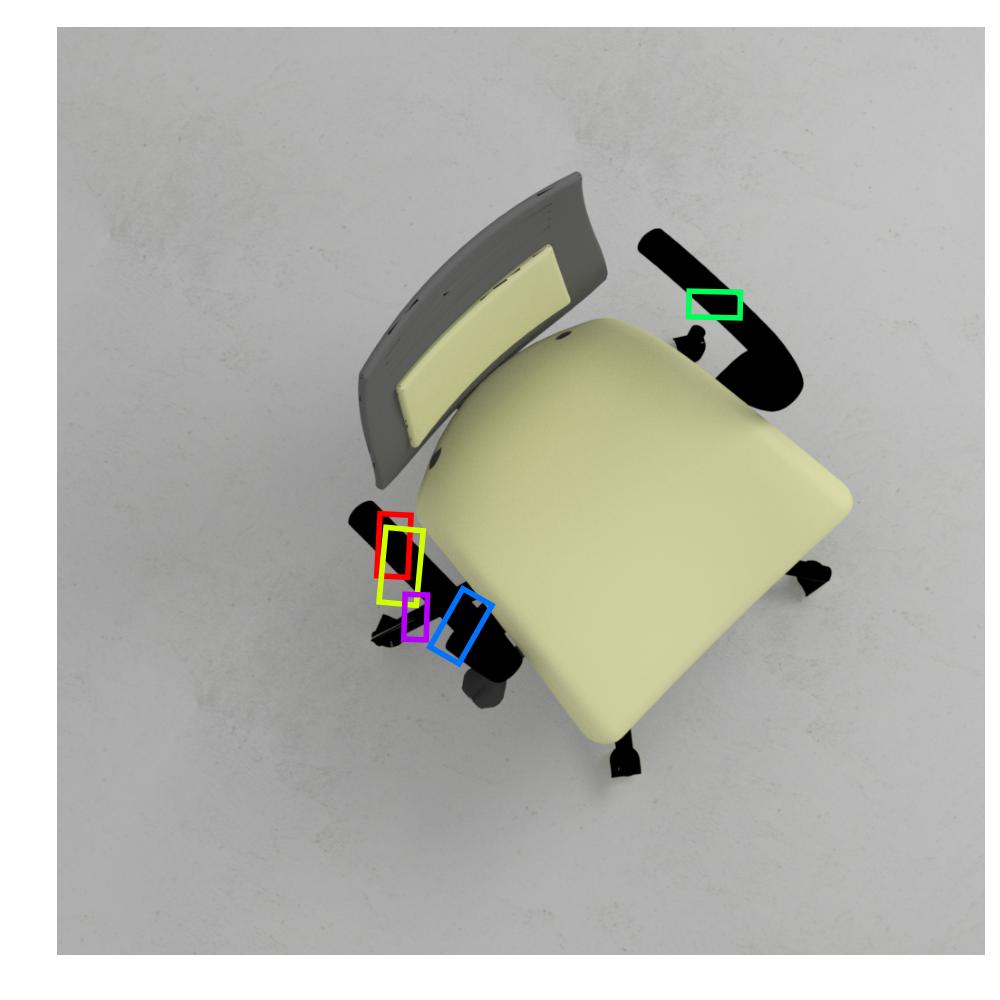}
        \end{overpic}
    \end{subfigure}
    \hfill
    \begin{subfigure}{\grenderwidth}
        \begin{overpic}[width=\textwidth,trim=100 10 10 0,clip]{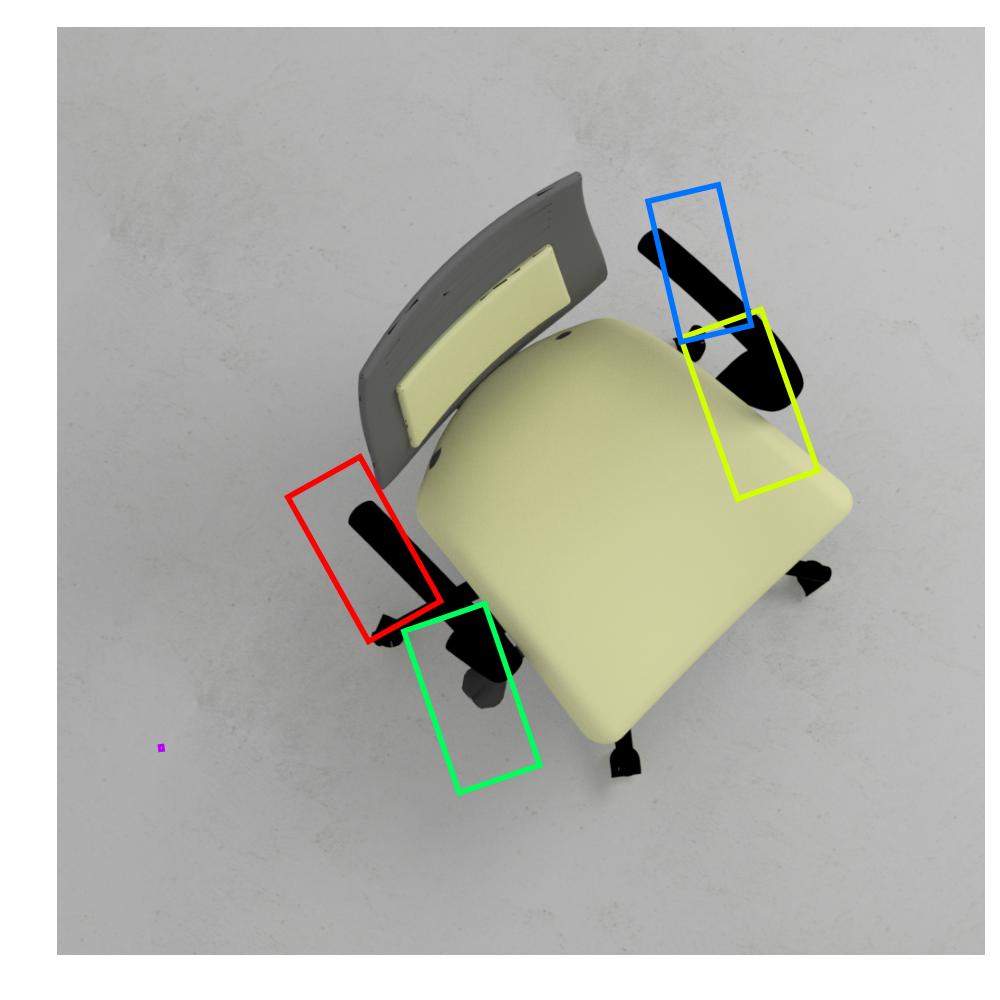}
        \end{overpic}
    \end{subfigure}
    \hfill
    \begin{subfigure}{\grenderwidth}
        \begin{overpic}[width=\textwidth,trim=100 10 10 0,clip]{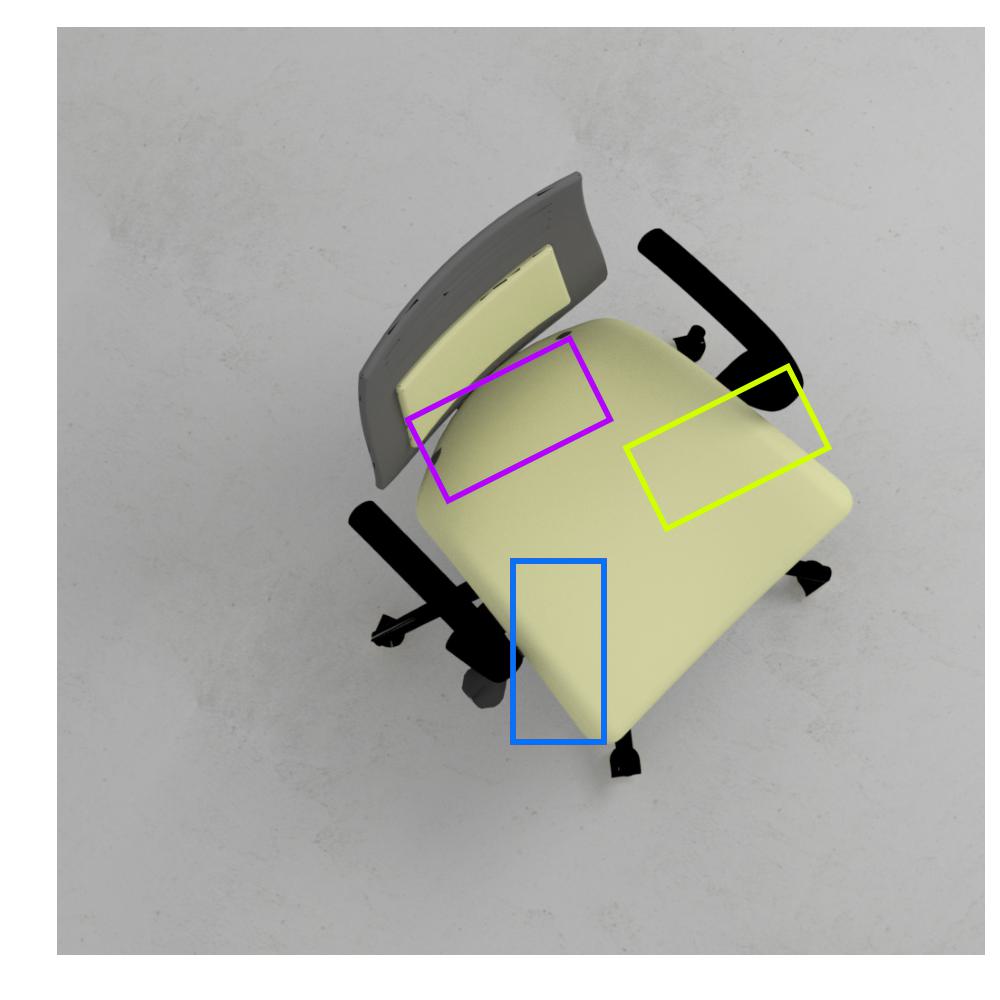}
        \end{overpic}
    \end{subfigure}
    \hfill
    \begin{subfigure}{\grenderwidth}
        \begin{overpic}[width=\textwidth,trim=100 10 10 0,clip]{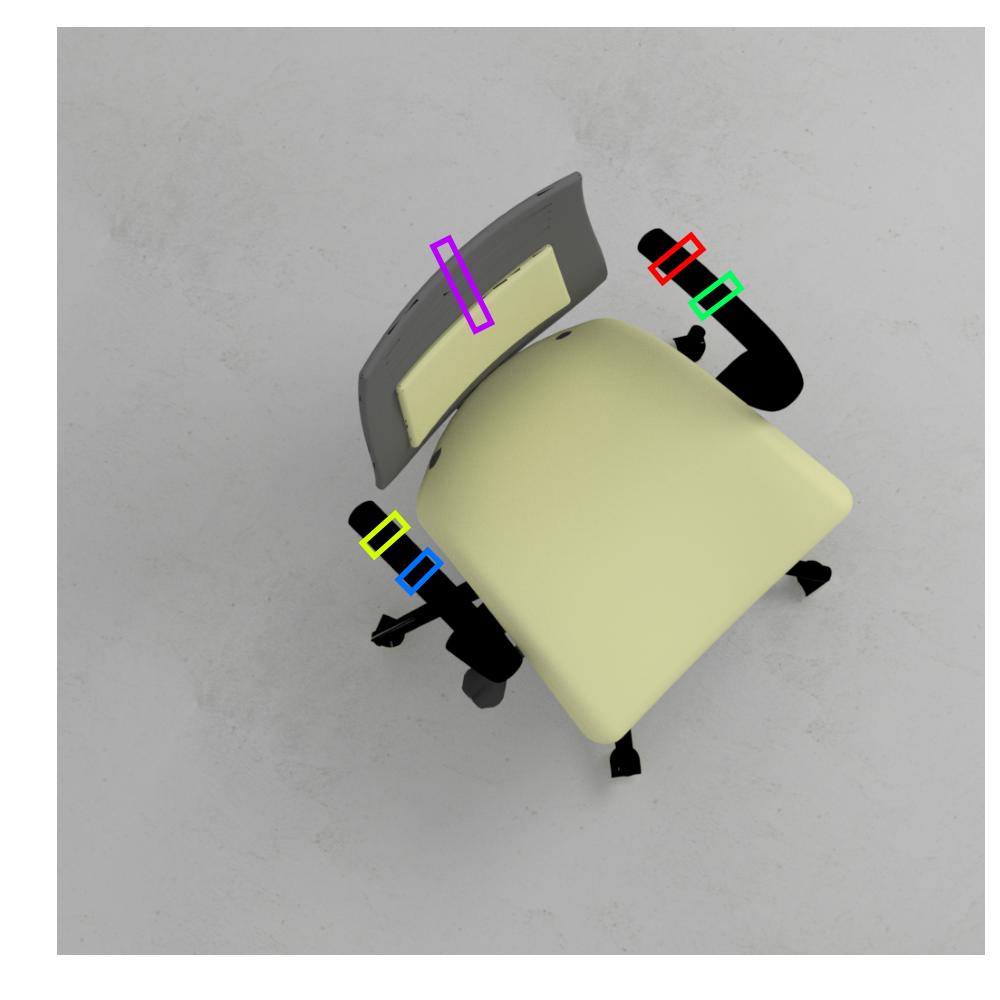}
        \end{overpic}
    \end{subfigure}
    \newline
    \begin{subfigure}{\grenderwidth}
        \begin{overpic}[width=\textwidth,trim=300 260 200 250,clip]{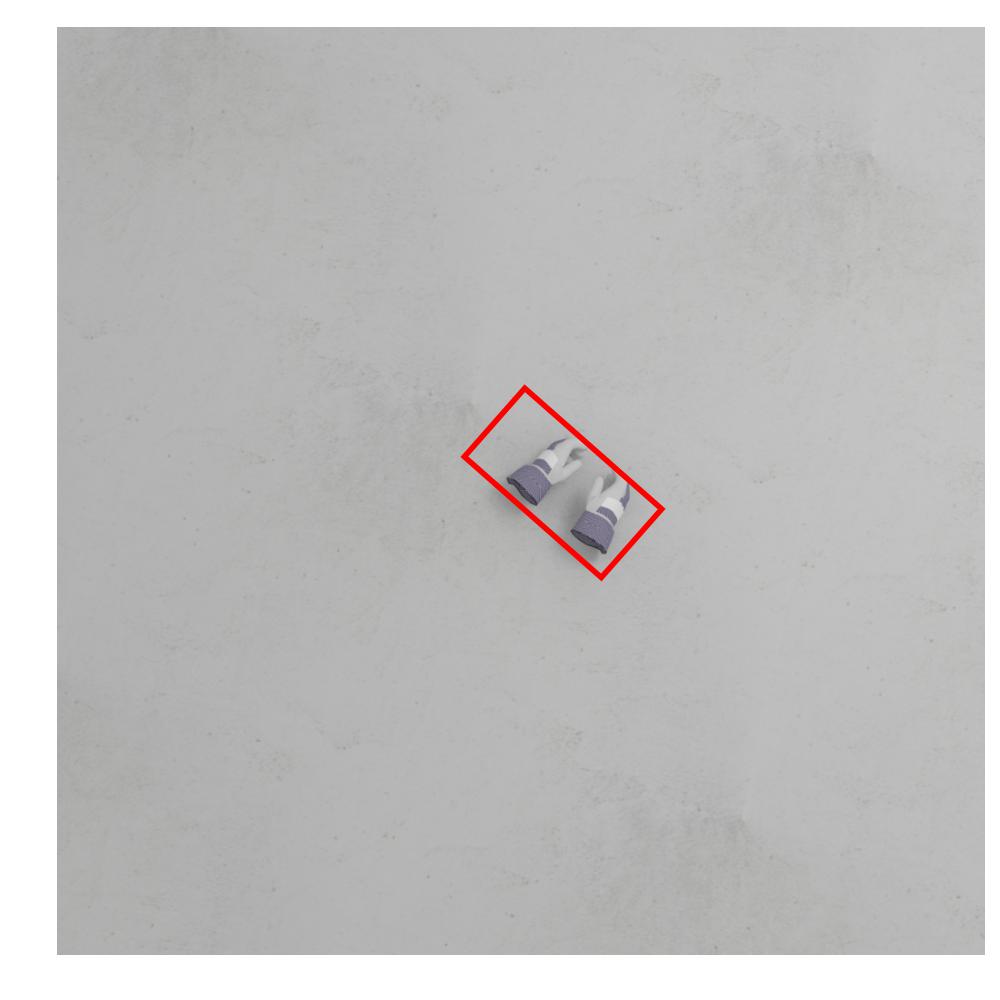}
        \end{overpic}
    \end{subfigure}
    \hfill
    \begin{subfigure}{\grenderwidth}
        \begin{overpic}[width=\textwidth,trim=300 260 200 250,clip]{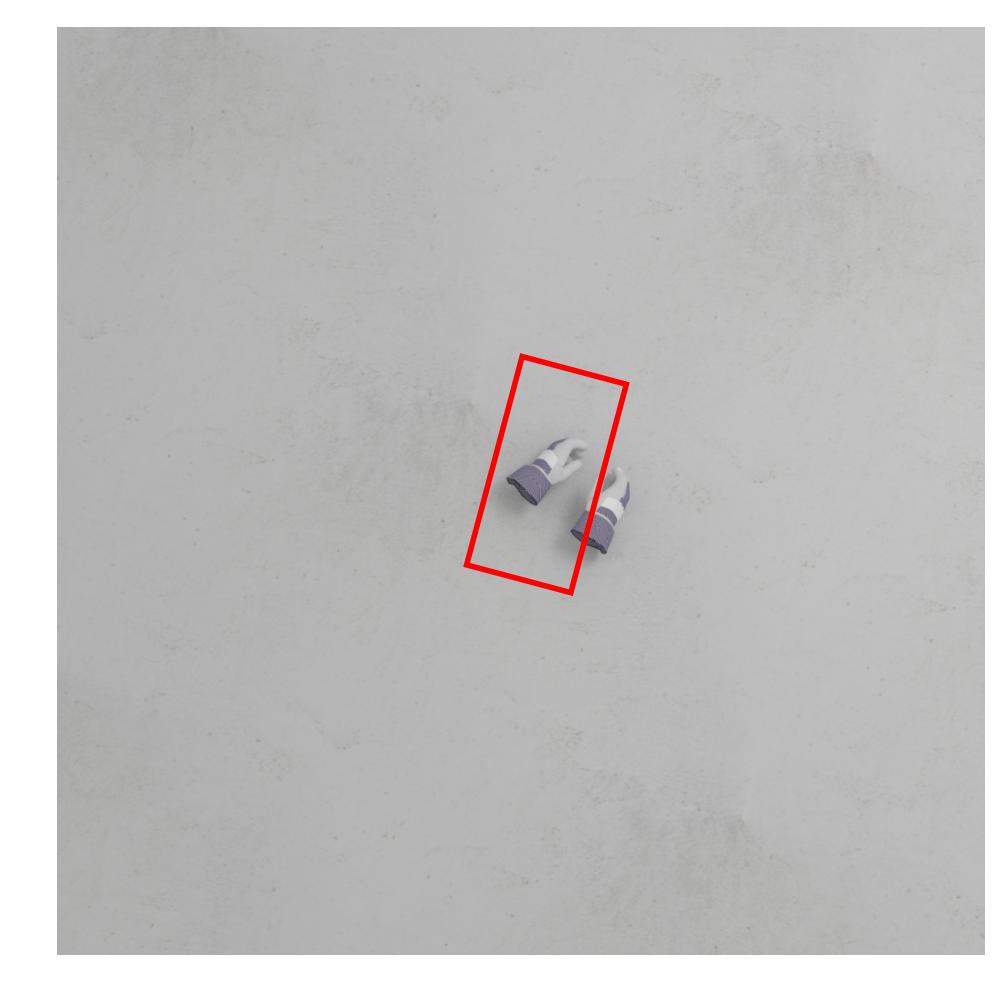}
        \end{overpic}
    \end{subfigure}
    \hfill
    \begin{subfigure}{\grenderwidth}
        \begin{overpic}[width=\textwidth,trim=300 260 200 250,clip]{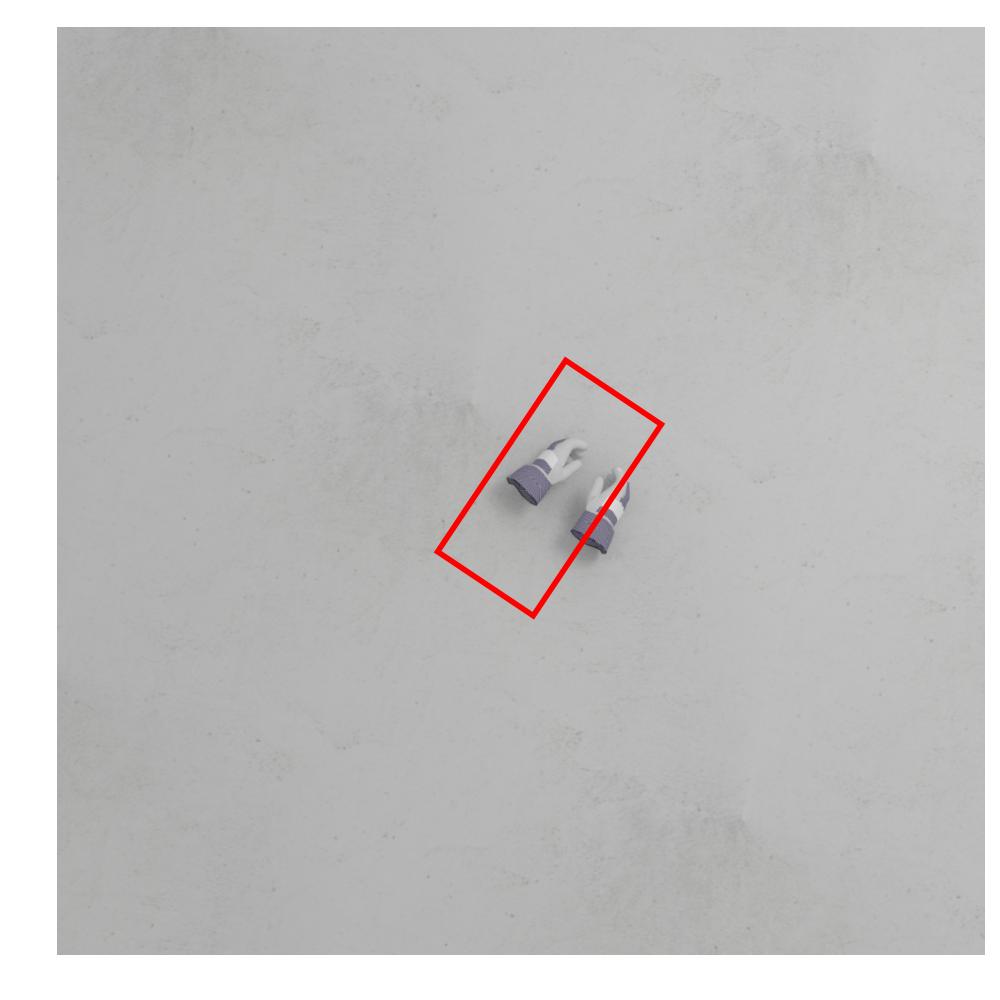}
        \end{overpic}
    \end{subfigure}
    \hfill
    \begin{subfigure}{\grenderwidth}
        \begin{overpic}[width=\textwidth,trim=300 260 200 250,clip]{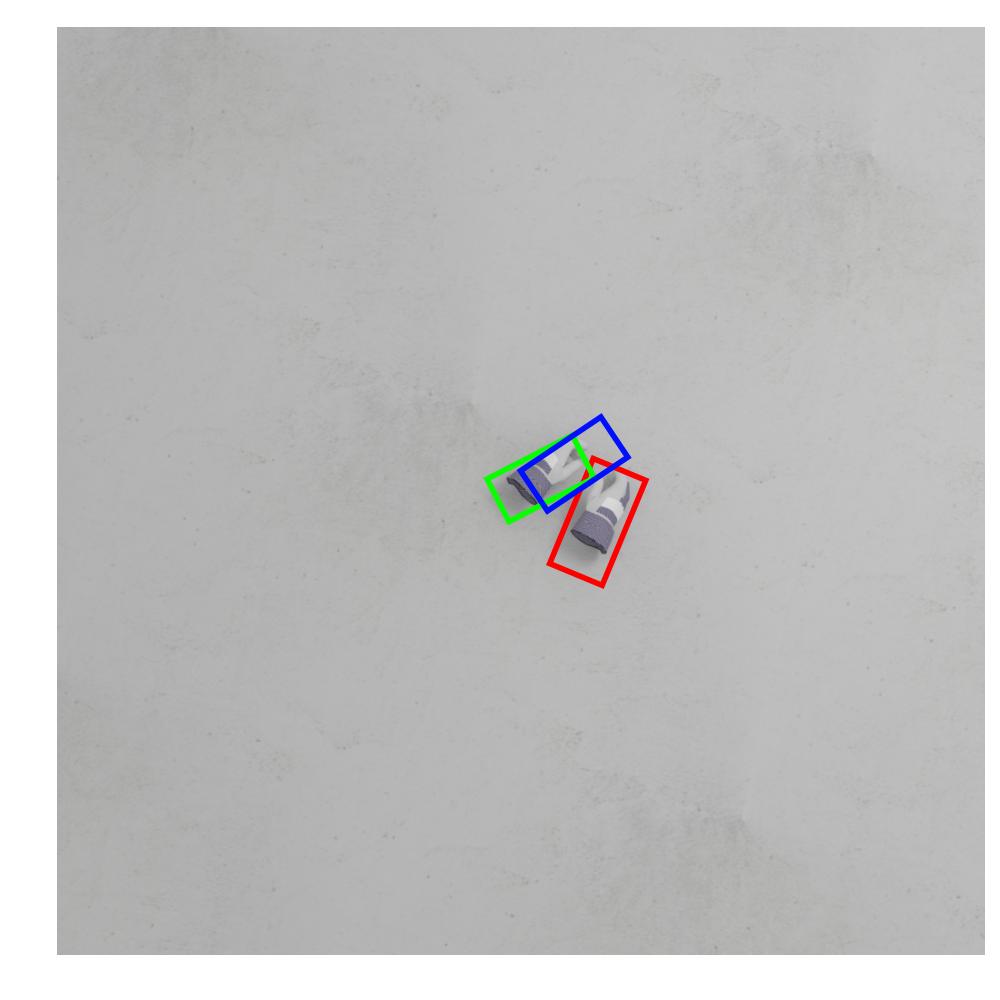}
        \end{overpic}
    \end{subfigure}
    \hfill
    \begin{subfigure}{\grenderwidth}
        \begin{overpic}[width=\textwidth,trim=300 260 200 250,clip]{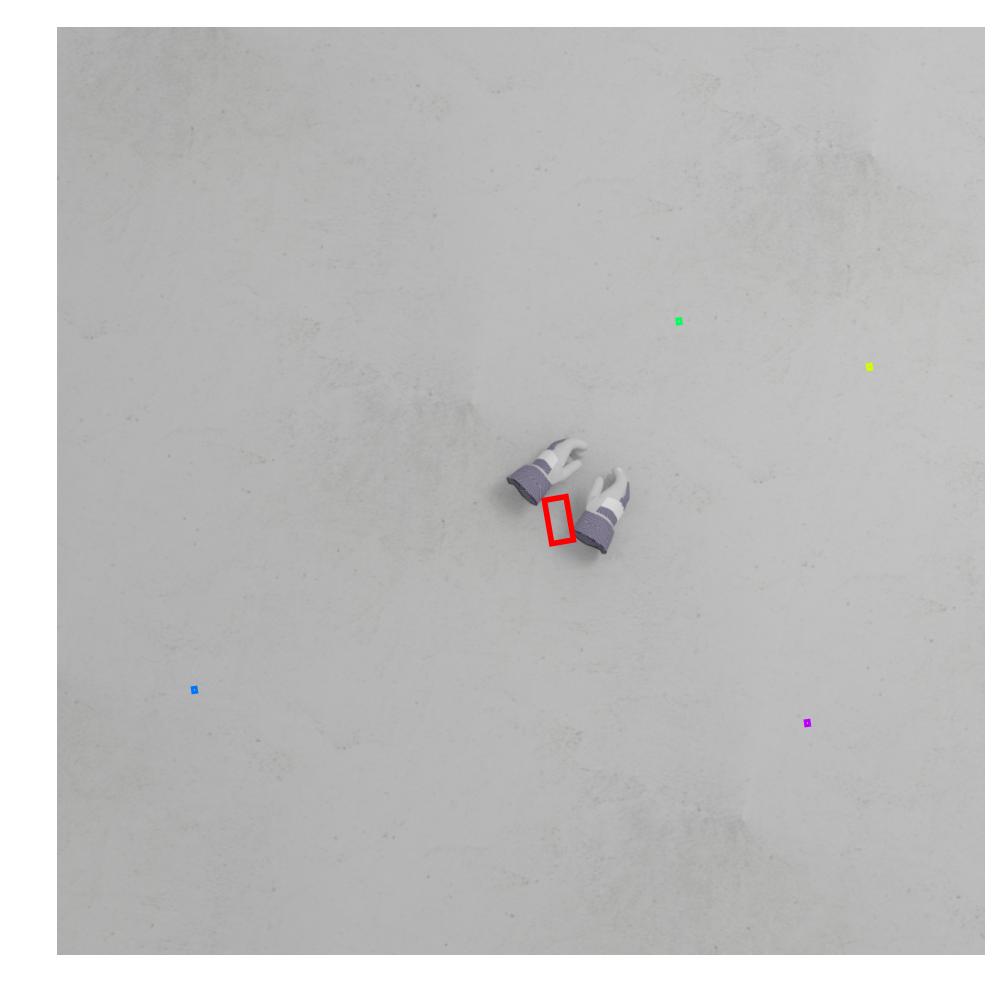}
        \end{overpic}
    \end{subfigure}
    \hfill
    \begin{subfigure}{\grenderwidth}
        \begin{overpic}[width=\textwidth,trim=300 260 200 250,clip]{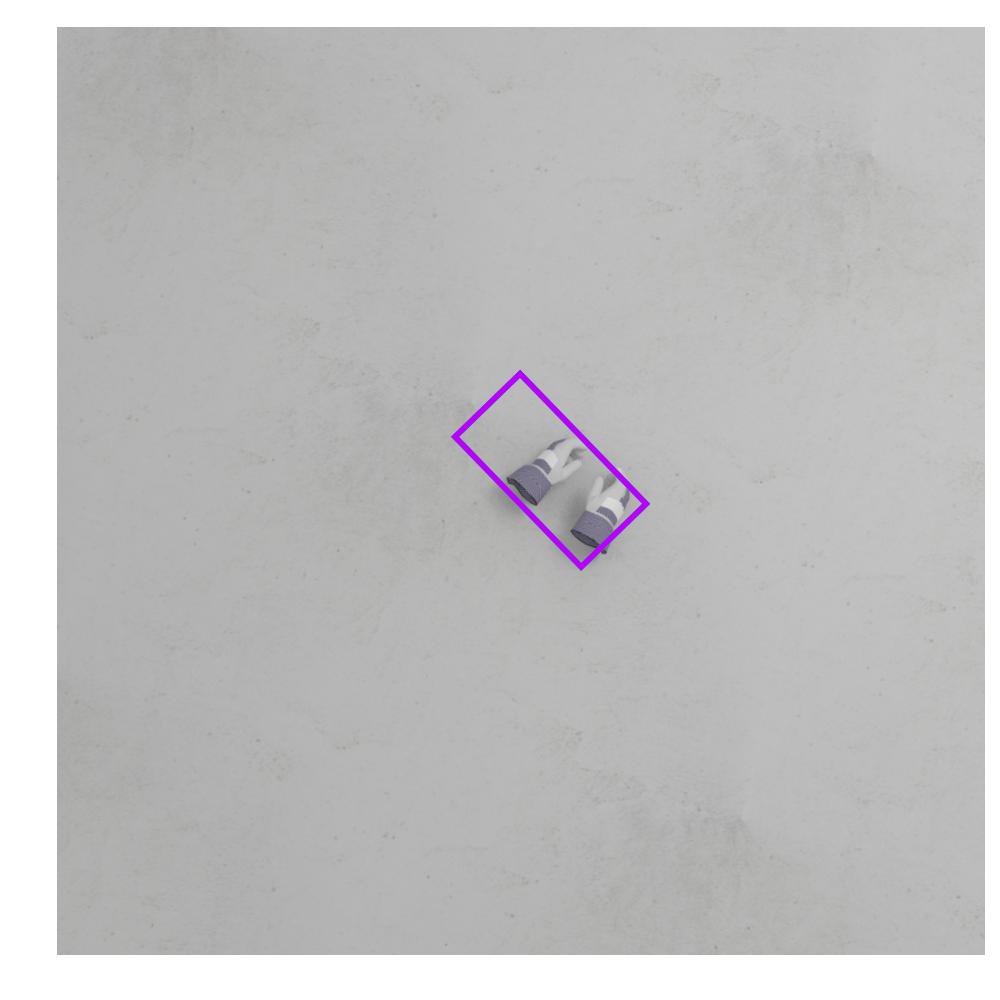}
        \end{overpic}
    \end{subfigure}
    \hfill
    \begin{subfigure}{\grenderwidth}
        \begin{overpic}[width=\textwidth,trim=300 260 200 250,clip]{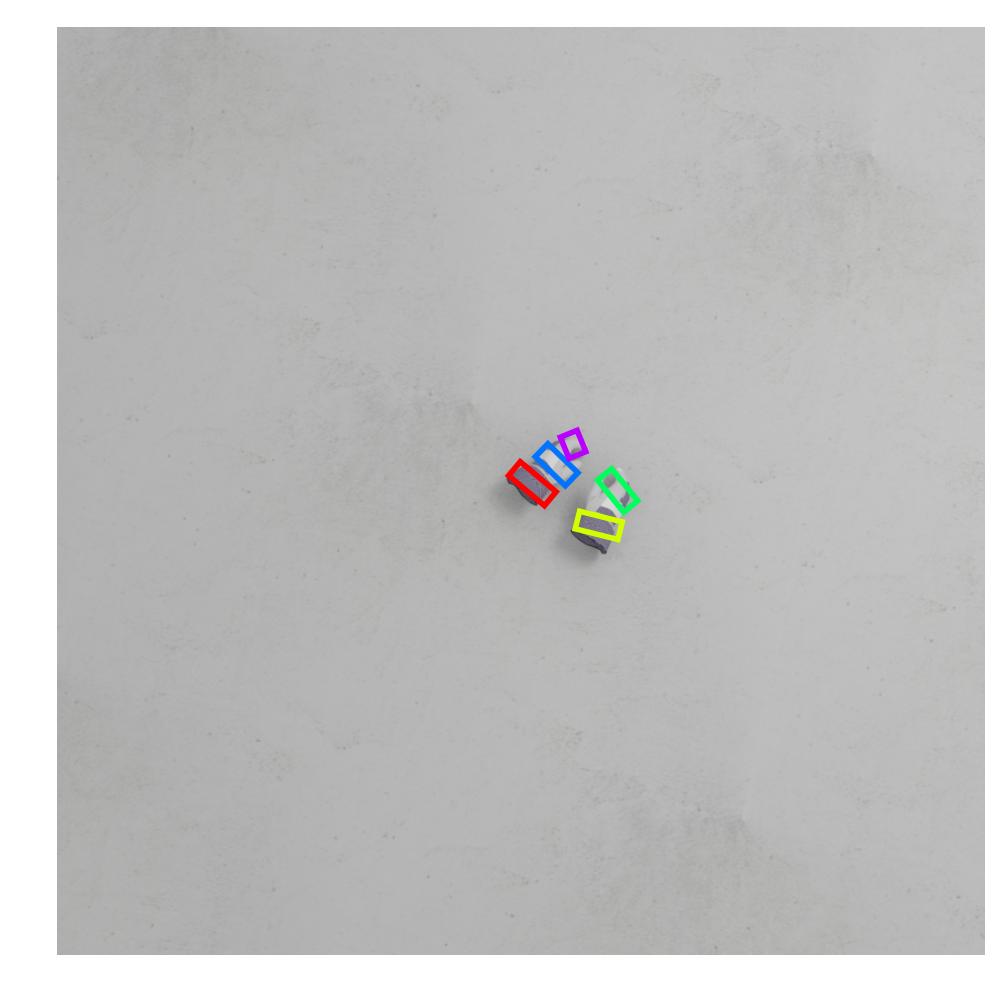}
        \end{overpic}
    \end{subfigure}
    \caption{Qualitative comparison of grasp detection on objects from the Jacquard dataset.
    For each method, we show five grasps per object, although for some methods the multiple outputs converge to a single grasp.
    Compared to baselines, our method detects more successful grasps across diverse object types.
    Prior methods often produce coarse or misaligned grasps, while our approach generates accurate and well-localized grasps that align with object geometry.
    }\label{fig:grasp-renders}\vspace{-3mm}
\end{figure*}

We evaluate our approach against state-of-the-art methods for grasp detection on images across the Cornell~\cite{jiang2011efficient} (real) and Jacquard~\cite{depierre2018jacquard} (simulated) datasets.
We use the open-source implementations of the following baseline methods for comparison:
GR-ConvNet \cite{kumra2020antipodal}, GG-CNN \cite{morrison2018closing}, SE-ResUNet \cite{ainetter2021end}, GraspSAM \cite{noh2025graspsam}, LGD \cite{vuong2024language}, and ShapeGrasp \cite{li2024shapegrasp}.
All baseline methods were trained on an NVIDIA RTX 5080 GPU system, except (1) LGD, where we use the authors' pretrained weights, and (2) ShapeGrasp, which is training-free.
Furthermore, we perform ablations to analyze the importance of the multi-step prompting structure, the effect of utilizing different VLMs (with our primary choice being the GPT-5 model \cite{hurst2024gpt}), and the choice of cross-domain object point cloud alignment.
Where unavailable, depth maps are generated by the ML Depth Pro model \cite{bochkovskii2024depth}, and segmentations are generated by the Segment Anything Model (SAM) \cite{kirillov2023segment}.
Finally, we demonstrate our approach for robotic picking in the real world. %

\subsection{Zero-Shot Grasping}

\begin{table}[t]
    \centering\vspace{2.5mm}
    \begin{tabular}{lcc}
    \toprule
    \multirow{2}{*}{Method} & \multicolumn{2}{c}{Success Rate (\%) $\uparrow$} \\
    \cmidrule(lr){2-3} & Cornell & Jacquard \\
    \midrule
    GR-ConvNet$^*$~\cite{kumra2020antipodal} & $72.14$~\scriptsize{$\pm 41.19$} & $59.62$~\scriptsize{$\pm 46.13$}\\
    GG-CNN$^*$~\cite{morrison2018closing} & $74.28$~\scriptsize{$\pm 40.30$} & $71.48$~\scriptsize{$\pm 42.76$}\\
    SE-ResUNet$^*$~\cite{ainetter2021end} &\underline{$86.07$}~\scriptsize{$\pm 28.65$} & $\mathbf{88.14}$~\textbf{\scriptsize{$\pm 30.43$}}\\
    GraspSAM$^*$~\cite{noh2025graspsam} & $67.50$~\scriptsize{$\pm 44.19$} & $73.71$~\scriptsize{$\pm 40.95$}\\
    \midrule
    LGD w/ Query$^{*\dagger}$~\cite{vuong2024language} & $37.98$~\scriptsize{$\pm 44.81$} & $24.40$~\scriptsize{$\pm 41.39$}\\
    LGD w/o Query$^{*\dagger}$~\cite{vuong2024language} & $32.26$~\scriptsize{$\pm 42.47$} & $24.40$~\scriptsize{$\pm 41.39$}\\
    ShapeGrasp w/ Depth$^\dagger$~\cite{li2024shapegrasp} & $47.14$~\scriptsize{$\pm 49.91$} & $54.32$~\scriptsize{$\pm 49.81$}\\
    ShapeGrasp w/o Depth$^\dagger$~\cite{li2024shapegrasp} & $47.14$~\scriptsize{$\pm 49.91$} & $57.55$~\scriptsize{$\pm 49.43$}\\
    ShapeGrasp w/ Oracle$^\dagger$~\cite{li2024shapegrasp} & $55.71$~\scriptsize{$\pm 49.67$} & $63.67$~\scriptsize{$\pm 48.09$}\\
    \midrule
    VLAD-Grasp$^\dagger$ (Ours) & $\mathbf{91.43}$~\textbf{\scriptsize{$\pm 28.00$}} & \underline{$85.43$}~\scriptsize{$\pm 36.15$}\\
    \bottomrule
    \end{tabular}
    \caption{\label{tab:metrics-ds}
        Comparison with baseline methods on the Cornell~\cite{jiang2011efficient} and Jacquard~\cite{depierre2018jacquard} datasets.
        Following prior work~\cite{jiang2011efficient}, a grasp is deemed successful if its associated rectangle has an Intersection-over-Union metric $\geq25\%$ with at least one ground truth annotation.
        Training on expert grasps is denoted by ($*$) and language-guidance is denoted by ($\dagger$).
        ShapeGrasp w/ Oracle denotes an upper bound on ShapeGrasp's confidence-based 
        depth heuristic~\cite{li2024shapegrasp}, reporting success if either the depth 
        or no-depth variant succeeds.
    }
    \vspace{-2mm}
\end{table}

Following the experimental setup from prior work~\cite{vuong2024grasp}, we evaluate the performance of our approach zero-shot on unseen object classes from the Cornell (70) and Jacquard (97) datasets.
We note that the grasp annotations in the Jacquard dataset are not fully exhaustive.
Hence, the IoU metric for grasp success will penalize valid proposals outside the ground truth's coverage (we discuss this in detail in Sec.~\ref{subsec:data_coherence}).
To account for this, we present results on objects for which sufficient ground truth coverage is provided.
All training-based baselines except LGD were trained on the seen subset and evaluated on the unseen subset of the same dataset.
For LGD, we use the author-provided weights pretrained on the GraspAnything++ dataset~\cite{vuong2024language} and evaluate zero-shot on both datasets with and without language-guided queries, following a similar ablation from the original paper (which boasts competitive zero-shot transfer performance).
However, we exclude evaluations on the GraspAnything++ dataset as the publicly available dataset contains inconsistencies between grasp annotations and object masks.

Results are presented in Tab.~\ref{tab:metrics-ds}.
On the Cornell dataset, VLAD-Grasp achieves superior performance than all baselines, leading by $\sim 5\%$.
On the Jacquard dataset, our method achieves competitive results at second place, trailing by $\sim3\%$.
It is important to note that while our method operates zero-shot on both datasets, all baselines apart from LGD and ShapeGrasp have been trained directly on the corresponding dataset.
All methods except SE-ResUNet, GraspSAM, and ShapeGrasp perform worse on the Jacquard dataset compared to the Cornell dataset, likely due to its higher object variety and relatively sparse grasp annotation distribution given object geometry.
For GraspSAM and ShapeGrasp, the performance difference may be attributed to their dependence on object masks, which are obtained from ground truth in simulation for the Jacquard dataset but are predicted from RGB images for the Cornell dataset.
We also notice that while query prompts help improve performance for LGD on the Cornell dataset by $\sim5\%$, they do not have a noticeable effect on the Jacquard dataset.

The methods most closely related to our language-driven, zero-shot transfer from large-scale pretraining are LGD and ShapeGrasp, which perform poorly on both the datasets.
We attribute LGD’s poor performance to the limited coverage of expert grasp annotations and the noisy ground truth in the Grasp-Anything++ dataset \cite{vuong2024grasp, vuong2024language} on which it was pre-trained.
ShapeGrasp performs convex decomposition of geometries to construct a decomposition graph, which is fed to an LLM to reason about spatial relations.
Hence reducing the rich visual context into textual description and employing heuristics to generate the object graph yields poor performance compared to our method which directly ingests vision-language information.

\subsection{Ablations}

To demonstrate the operating principle of VLAD-Grasp and its design choices, we provide ablations for our approach.
We first ablate on the choice of VLM, in addition to examining the importance of the three-step prompting procedure (Sec.~\ref{subsec:gen}).
We then examine the effect of different point cloud alignment methods for the cross-domain object alignment, comparing with our method comprising principal component-based correspondence-free optimization (Sec.~\ref {subsec:align}).

\begin{table}[t]
    \centering\vspace{2.5mm}
    \begin{tabular}{lcccc}
    \toprule
    \multirow{2}{*}{VLM} & \multirow{2}{*}{SR (\%) $\uparrow$} & 
    \multicolumn{2}{c}{\# Tokens Used $(\times10^3)$} \\ 
    \cmidrule(lr){3-4} &  & Output & Reasoning \\
    \midrule
    GPT 5~\cite{hurst2024gpt} & $91.43$~\scriptsize{$\pm 27.99$} & $1.69$~\scriptsize{$\pm 1.01$} & $2.11$~\scriptsize{$\pm 0.56$} \\
    GPT 5 w/o 3-Step & $78.57$~\scriptsize{$\pm 41.03$} & $1.15$~\scriptsize{$\pm 0.52$} & 0 \\
    Gemini 2.5~\cite{comanici2025gemini} & $84.29$~\scriptsize{$\pm 36.39$} & $1.73$~\scriptsize{$\pm 0.15$} & $2.54$~\scriptsize{$\pm 1.15$} \\
    \bottomrule
    \end{tabular}
    \caption{Success Rate (SR), and number of output and reasoning tokens used for GPT 5 (with and without 3-Step prompting) and Gemini 2.5.}
    \label{tab:ablation_vlm}
    \vspace{-3mm}
\end{table}

Shown in Tab.~\ref{tab:ablation_vlm}, GPT-5 performs the best in terms of generating a feasible grasp, with a success rate of $91.43\%$.
Without the 3-Step prompting procedure, the success rate drops by more than $12\%$, however this trades off performance with cost and consistency, using $\sim 32\%$ fewer tokens as well as lower variance in grasp success.
This aligns with our empirical observations that both the intermediate text-based reasoning $R^g$ and the specificity of the generated query prompt $T^g_2$ helps guide the image generation process to obtain a better goal image $I_G$ and, consequently, a better object grasp.
Gemini 2.5~\cite{team2023gemini} performs $\sim7\%$ worse despite the higher number of reasoning tokens used, although with lower variance in grasp success.
We also explored utilizing a combination of Llama 4~\cite{touvron2023llama} for text-based reasoning and StableDiffusion's image-to-image model~\cite{rombach2022high} for generating goal images.
While the text outputs are reasonable, the generative model had extremely poor prompt adherence on the dataset images, and hence, we exclude it from Tab.~\ref{tab:ablation_vlm}.

\begin{table}[t]
    \centering\vspace{2.5mm}
    \begin{tabular}{lccc}
    \toprule
    \multirow{2}{*}{Method} & \multirow{2}{*}{SR (\%) $\uparrow$} & \multicolumn{2}{c}{Alignment Metric $\downarrow$} \\ 
    \cmidrule(lr){3-4}
     &  & CD $(10^{-3})$ & UHD $(10^{-2})$ \\
    \midrule
    TEASER++~\cite{yang2020teaser} & $55.71$~\scriptsize{$\pm 49.67$} & $0.24$~\scriptsize{$\pm 0.40$} & $5.68$~\scriptsize{$\pm 3.91$} \\
    ICP~\cite{rusinkiewicz2001efficient} & $34.29$~\scriptsize{$\pm 47.47$} & $76.8$~\scriptsize{$\pm 145.1$} & $50.8$~\scriptsize{$\pm 68.8$} \\
    PCA+Opt (Ours) & $\mathbf{91.43}$~\scriptsize{$\pm 27.99$} & $0.17$~\scriptsize{$\pm 0.17$} & $7.21$~\scriptsize{$\pm 6.48$}  \\
    \bottomrule
    \end{tabular}
    \caption{\label{tab:ablation_align}
    Success Rate (SR), Chamfer Distance (CD), and Unidirectional Hausdorff Distance (UHD) for TEASER++, Iterative Closest Point (ICP), and principal component alignment followed by correspondence-free optimization (PCAling+CfOpt; Ours).
    }\vspace{-2mm}
\end{table}

We present results for point cloud alignment approaches in Tab.~\ref{tab:ablation_align}.
In terms of grasp success rate, our approach (PCAlign + CfOpt) significantly outperforms the comparison methods -- TEASER++~\cite{yang2020teaser} by about 36\% and Iterative Closest Point (ICP)~\cite{rusinkiewicz2001efficient} by about 57\% -- with a lower variance in grasp success rate.
ICP performs the worst in terms of the Chamfer Distance (CD) and Unidirectional Hausdorff Distance (UHD), which we attribute to its myopic optimization routine which easily gets trapped in a local minima.
Lastly, we notice that our method achieves $\sim 30\%$ lower CD but with a $\sim 30\%$ higher UHD, likely due to the choice of CD as the correspondence-free loss $\mathcal{L}$ for optimization.
This tradeoff reflects a key strength of our approach:
Unlike methods that are sensitive to local feature alignment, our optimization prioritizes the global shape structure.
This means that, for a case like ours where local features between the input object $o_S$ shape and generated object $o_G$ shape may vary or where the depth prediction errors distort finer details, the overall shape remains sufficiently similar across the two object pointclouds $P^o_S, P^o_G$ and are preserved to support downstream tasks. 

\subsection{Real World Deployment}

We evaluate VLAD-Grasp on real-world robotic grasping to demonstrate its feasibility for realistic tasks.
We deploy on a Franka Research 3 robot with a Franka Panda Hand comprising a parallel jaw gripper.
We obtain RGB-D observations with a wrist-mounted ORBBEC Femto Mega camera, from which VLAD-Grasp outputs a grasp candidate in the image space.
The robot then uses its end-effector pose and the camera's relative extrinsics to obtain an approximate grasp pose above the object.
Next, the robot hand approaches the object, closes the gripper to grasp the object, and lifts for several seconds to validate a successful hold.
Results in Tab.~\ref{tab:metrics-rw} demonstrate the efficacy of VLAD-Grasp in generating consistent, physically plausible grasps for a variety of real-world objects.
We complement the results with videos demonstrating the robustness of our method's performance in the real-world (see the supplementary material).

\begin{table}[t]
    \centering\vspace{2.5mm}
    \begin{tabular}{lccccc}
    \toprule
    \multirow{2}{*}{Method} & \multicolumn{5}{c}{SR $(\cdot/10$) $\uparrow$} \\ \cmidrule(lr){2-6}
                            & Bottle & Cup & Spatula & VRCtrl. & Drone \\
    \midrule
    GR-ConvNet~\cite{kumra2020antipodal} & 7/10 & 4/10 & 3/10 & 8/10 & 10/10 \\
    GG-CNN~\cite{morrison2018closing} & 5/10 & 5/10 & 5/10 & 7/10 & 10/10 \\
    SE-ResUNet~\cite{ainetter2021end} & 6/10 & 4/10 & 3/10 & 5/10 & 5/10 \\
    {GraspSAM}~\cite{noh2025graspsam} & 2/10 & 2/10 & 3/10 & 4/10 & 6/10 \\
    LGD w/ Query~\cite{vuong2024language} & 4/10 & 2/10 & 1/10 & 8/10 & 8/10 \\
    {ShapeGrasp}~\cite{li2024shapegrasp} & 10/10 & 7/10 & 8/10 & 4/10 & 6/10 \\
    \midrule
    VLAD-Grasp (Ours) & 10/10 & 8/10 & 8/10 & 9/10 & 10/10 \\
    \bottomrule
    \end{tabular}
    \caption{Grasping objects in the real-world.
        We evaluate each method on 10 different orientations of 5 different objects and report the number of successes.
        Our approach outperforms or matches baselines across all objects.
    }\label{tab:metrics-rw}
\end{table}

\section{Discussion}

\subsection{Dataset Coherence}\label{subsec:data_coherence}
As the scale and object diversity of datasets increase, it becomes challenging to provide fully exhaustive grasp annotations for each object geometry.
For both the Cornell and Jacquard datasets, we observe that the distribution of ground-truth grasp annotations lack full coverage of ground-truth grasp modes, given the object geometries. 
Grasp annotations for the Cornell dataset sometimes miss viable grasp poses.
However, this is mitigated by the lower object diversity and the use of human annotation for labeling ground-truth grasps.
For the Jacquard dataset, while we are not privy to the sampling heuristic used, the annotations for successful grasps vary in quality, number, and distribution across the dataset.
Naturally, this affects methods trained on these datasets by biasing them to the specific distribution of grasps seen in the training set, which upper-bounds their performance. %

\begin{figure}[t]
    \centering\vspace{2mm}
    \begin{subfigure}{0.48\linewidth}
        \begin{overpic}[width=\textwidth,trim=280 225 200 180, clip]{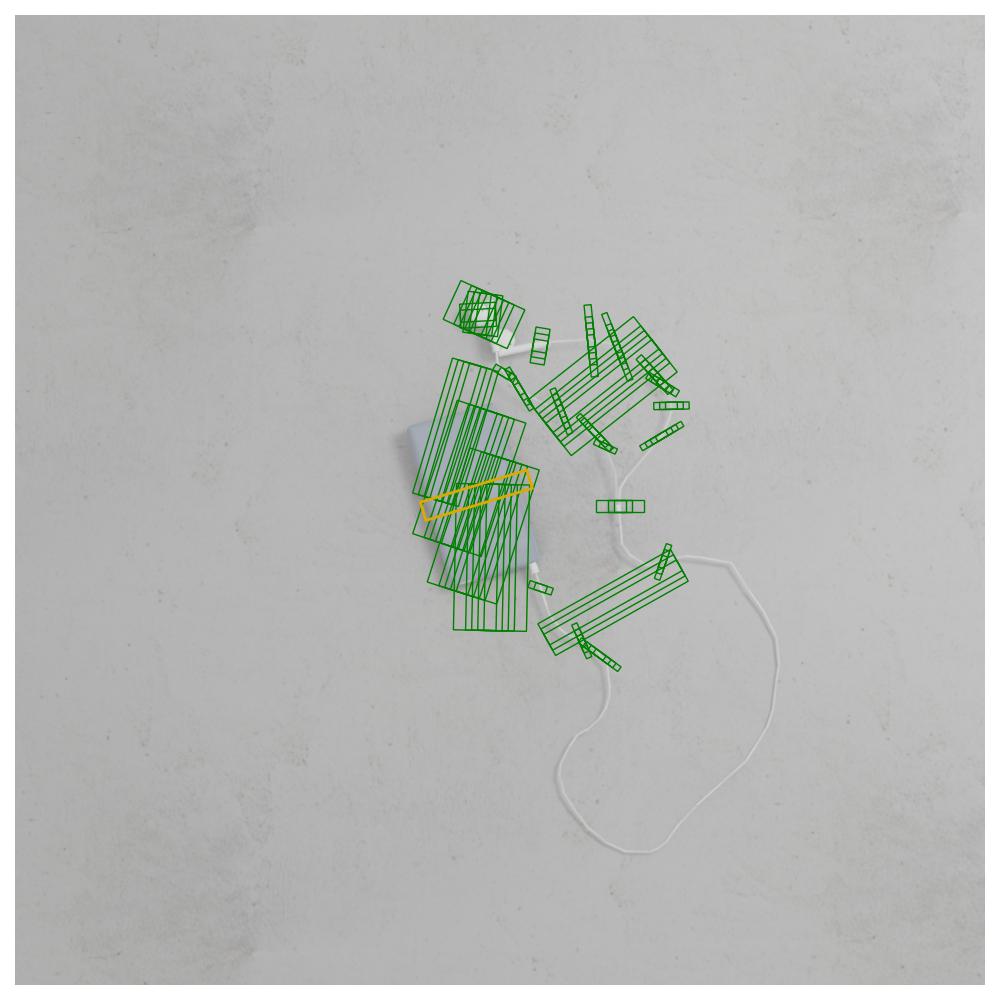}
        \end{overpic}
    \end{subfigure}
    \begin{subfigure}{0.48\linewidth}
        \begin{overpic}[width=\textwidth,trim=200 100 120 95,clip]{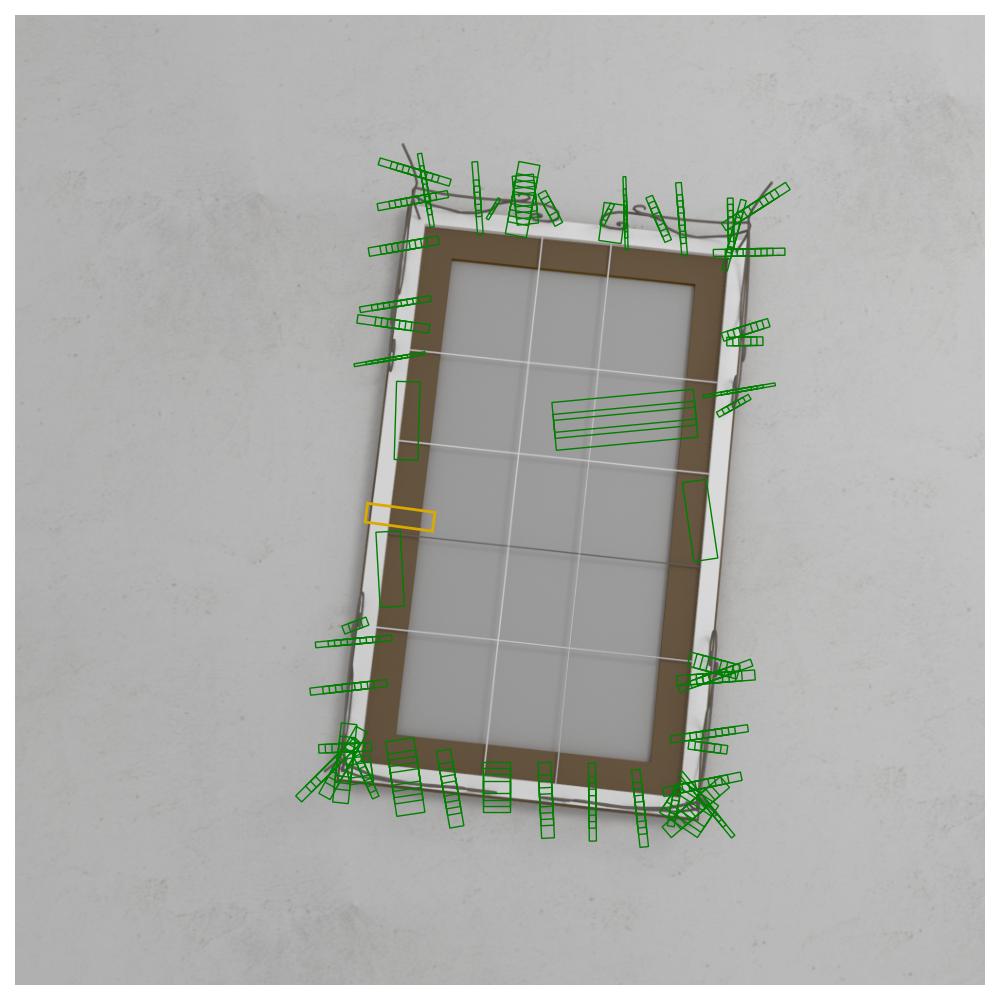}
        \end{overpic}
    \end{subfigure}
    
    \caption{Examples of viable grasps generated by our approach (yellow), improperly marked as failures due to high angular mismatch (\emph{left}) and missing ground truth annotations near the predicted grasp (\emph{right}).
}\label{fig:dataset_bad}
\end{figure}

Our method does not suffer from such issues, as it does not learn to grasp directly.
Rather it reasons about object geometry and visual spatial relationships to generate a grasp.
Therefore, our approach can generate valid grasps that are not represented in the evaluation dataset.
However, some grasps generated by our method are incorrectly marked as failures, as indicated in Fig.~\ref{fig:dataset_bad}.
Our approach proposes grasping the music player (\emph{left}) along the short axis, which corresponds to insufficient IoU with ground truth annotations, and grasping the canvas (\emph{right}) with its back facing the camera, with zero overlap with ground truth annotations.
Hence, we argue that the performance indicated by training-free approaches in Tab.~\ref{tab:metrics-ds} is an underestimate of the potential performance in practice (as demonstrated in Tab.~\ref{tab:metrics-rw}), and therefore, better grasp evaluation methods and metrics need to be devised.

\subsection{Limitations}
While the large-scale training of VLMs allows them to perform and generalize well for most generative tasks, this is not always reliable.
In some cases, the VLM fails to accurately interpret the object in the input image, resulting in surreal or distorted structures that disrupt the global geometry of the original object.
Such failures may arise from imperfect object understanding, the intrinsic instability in the generation process, or the fact that inferring 3D information from a single 2D observation is an ill-posed problem.
Similarly, prompt adherence is not guaranteed, and the VLM may generate content that deviates from the specified instructions, resulting in outputs that cannot be plausibly translated into actionable grasps.
However, as the state of research progresses, better versions can be directly plugged into our approach for superior performance with better efficiency and throughput.

\section{Conclusion}
We propose VLAD-Grasp, a zero-shot grasp synthesis framework that leverages the reasoning and generative capabilities of large-scale vision–language models.
By formulating grasp detection as the generation of a virtual cylindrical proxy intersecting the object's geometry (encoding an antipodal grasp axis in image space), and subsequently lifting this representation into 3D through depth prediction and geometric alignment, our approach bypasses the need for expert grasp annotations or retraining.
Our training-free pipeline demonstrates that foundation models can serve as powerful priors for robotic manipulation.

Our experiments highlight several key strengths of VLAD-Grasp.
First, the method achieves competitive or superior grasp success rates compared to state-of-the-art supervised models, despite no task-specific training.
Second, the integration of structured prompting and geometric consistency ensures robust transfer from virtual grasping concept images to executable grasps in the real world.
Finally, the successful deployment on a Franka Research 3 robot reinforces that the approach generalizes to novel, previously unseen objects, underscoring the generalizable reasoning capabilities of VLMs for robotic grasping.

\section*{Acknowledgments}
We acknowledge the assistance of LLMs (ChatGPT, Claude) towards refining minor parts of the manuscript text.

\renewcommand*{\bibfont}{\footnotesize}
\bibliographystyle{IEEEtranN}
\bibliography{IEEEabrv,root}

\end{document}